\documentclass{article}
\usepackage{amssymb}
\usepackage{amsmath,amsfonts}
\usepackage{array}
\usepackage{booktabs}
\usepackage{textcomp}
\usepackage{stfloats}
\usepackage{url}
\usepackage{verbatim}
\usepackage{graphicx}
\usepackage{capt-of}
\usepackage{cite} 
\usepackage{xcolor}
\usepackage{hyperref}
\usepackage{float}
\usepackage[export]{adjustbox}
\usepackage{booktabs}
\usepackage[ruled]{algorithm2e}
\usepackage{algpseudocode}
\usepackage{dsfont}
\usepackage{amsthm}
\usepackage{etoolbox}
\usepackage{microtype}
\usepackage{bbm}
\usepackage{wrapfig}
\usepackage{multicol}
\usepackage[nolist,nohyperlinks]{acronym}
\usepackage{pifont}
\newcommand{\cmark}{\ding{51}}
\newcommand{\xmark}{\ding{55}}
\providetoggle{anon}

\SetKw{Params}{Params}
\DontPrintSemicolon
\usepackage[caption=false,font=normalsize,labelfont=sf,textfont=sf]{subfig}
\usepackage[preprint]{corl_2026} 

\usepackage{tikz}
\usepackage{accsupp}
\newcommand{\PushTsymbol}{%
  \begingroup
  \setbox0=\hbox{T}%
  \rlap{\textcolor{black!0}{T}}%
  \makebox[\wd0][c]{%
    \tikz[baseline=0.1ex,line width=0.06em,scale=0.8]{
      \draw (-0.14em,0) rectangle (0.14em,0.7em);
      \draw (-0.49em,0.7em) rectangle (0.49em,0.98em);
    }%
  }%
  \endgroup
}

\begin{acronym}
  \acro{IL}{Imitation Learning}
  \acro{DAgger}{Dataset Aggregation}
\end{acronym}

\title{Training-Free Imitation Learning with\\ Closed-Form Diffusion Policies}

\author{
  Raghav Mishra, Ian R. Manchester\\
  Australian Center for Robotics, ARIAM Hub,\\
  and School of Aerospace, Mechanical and Mechatronic Engineering \\
  University of Sydney \\
  \texttt{\{raghav.mishra, ian.manchester\}@sydney.edu.au} \\
}

\begin{document}
\maketitle
\begin{center}
\includegraphics[width=0.9\textwidth]{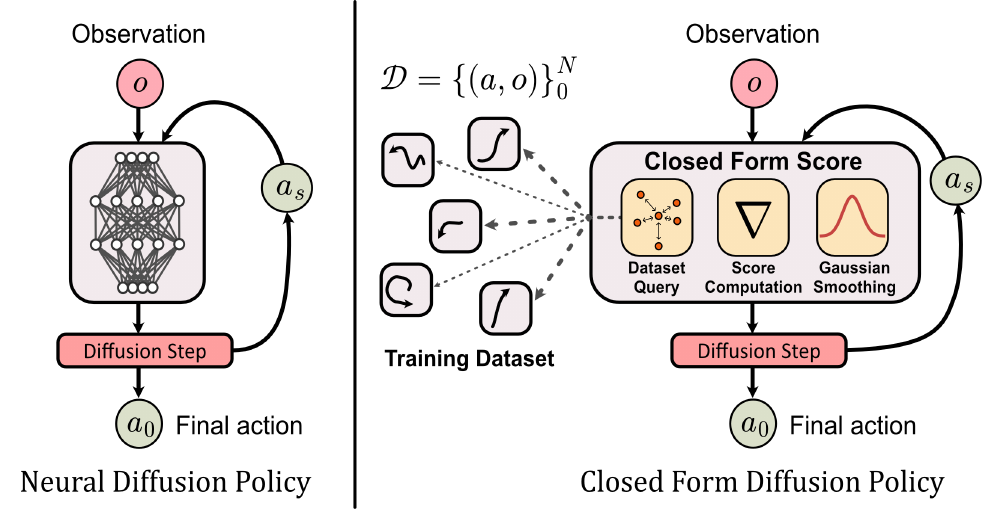}
\end{center}

\begin{abstract}
While diffusion-based policies have impressive performance and expressivity, their long offline training slows down the data collection and policy deployment loop. We introduce Closed-Form Diffusion Policies, a class of training-free diffusion-based policies for imitation learning using the closed-form score derived from the demonstration dataset. We deploy CFDP with real-time inference with a mobile CPU in hardware experiments, showing it can successfully perform imitation directly from the dataset in milliseconds and with faster inference than neural diffusion policies. In experiments on imitation learning benchmarks, we show that CFDP is competitive against neural baselines that require hours of training, providing a favorable tradeoff between training time and performance. Finally, we show how closed-form diffusion policies act as a composable primitive that enables data-driven inference-time editing of pre-trained neural diffusion policies, including policy guidance and novel demonstration augmentation.
\end{abstract}


\section{Introduction}
\label{sec:intro}
Imitation learning (IL) enables robots to perform tasks that are difficult to specify explicitly by learning from expert demonstrations. However, in practice it is often unclear whether a collected dataset would be sufficient to train a suitable policy until after deployment. Consequently, IL is frequently performed in an iterative manner, alternating between data collection and policy retraining \cite{ross2011reduction}. This process can be particularly costly in domains such as underwater or space robotics, where collecting additional demonstrations requires significant time, expense, and operational effort.

Recent advances in robot learning have increasingly framed imitation learning as a generative modeling problem, motivated in part by the multi-modal distribution of expert actions. Diffusion models have emerged as a particularly successful generative architecture, demonstrating strong performance in visuomotor policy learning and the imitation of multi-modal expert trajectory distributions~\cite{chi_diffusion_2023}. However, diffusion policies are computationally expensive to train, often requiring hours or days of offline optimization before they can be deployed. This high training cost arises because the score network that underlies their generation must be learned across many noise scales and noising trajectories. As a result, diffusion policies exacerbate the cost of learning policies with IL, since each additional deployment cycle may require substantial retraining time alongside expensive data collection.

\paragraph{Contributions.}
Motivated by these challenges, we introduce the following contributions:
\begin{itemize}
 \item We introduce \textbf{Closed-Form Diffusion Policies (CFDPs)}, a class of training-free diffusion policies for imitation learning based on conditional closed-form score functions.
 \item We demonstrate that CFDPs can be deployed with a mobile CPU to control a robot in \textbf{milliseconds without any training} with 7x faster inference than neural diffusion policies. 
 \item We show experimental results that demonstrate that CFDPs \textbf{perform competitively to neural policies} on low-dimensional imitation learning tasks. 
 \item We show CFDPs can perform \textbf{inference-time policy editing} on pre-trained neural diffusion policies, including closed-form guidance and novel demonstration augmentation.
\end{itemize}

\section{Related Work}
\subsection{Imitation Learning}
\ac{IL} enables robots to learn tasks by mimicking expert demonstrations, typically using behavior cloning. While behavior cloning has been widely applied, it struggles with compounding errors where discrepancies between the demonstration observations and the true observations leads to failure. To address these limitations, many approaches have been proposed. Online learning methods like DAgger~\cite{gordon2011_dagger} mitigate this by iteratively collecting new expert data during training. Recently, policy classes such as Diffusion Policies~\cite{chi_diffusion_2023} and Action Chunking Transformers~\cite{Zhao_ACT} have shown impressive performance over a wide variety of problems, with common policy architectures features like action chunking, generative model backbones conditioned on observation history, etc. Recent theoretical analysis has started to validate the benefits of architectural choices such as action chunking in common IL methods~\cite{zhang_action_theory}.

\subsection{Diffusion-based Policies}
Diffusion models~\cite{sohl-dickstein_deep_2015} have shown to be an expressive class of policies. Diffusion policies or planners~\cite{chi_diffusion_2023, janner_planning_2022} can be trained on human demonstration data, or can distill solutions from trajectory optimization~\cite{li25_diffusolve} or model predictive control~\cite{huang25_diffusion_learns_mpc} data collected offline. Additionally, diffusion models can be an expressive policy class for multi-modal actions in reinforcement learning~\cite{wang2022diffusionRL}. In contrast to our training-free approach, we call these methods ``neural diffusion policies'' (NDPs). In robotics, there has been a large focus on NDPs, especially as they can be guided at inference to reduce user-specified cost~\cite{carvalho_motion_2024}, incorporate constraints~\cite{li24_constraintawarediffusion}, or steer policies towards desired actions~\cite{du2025dynaguidesteeringdiffusionpolices}.

In parallel to learning-focused approaches, connections between diffusion and optimization have prompted sampling-based optimization methods with diffusion-inspired noising schedules~\cite{pan_model-based_2024, xue2025full, mishra2026_ebmbd} to benefit from the global optimization properties of Gaussian continuations~\cite{nesterov_random_2017}. However these methods generally don't leverage prior data and require dynamics models and cost functions.

\section{Technical Background}

\subsection{Diffusion Models}
Diffusion models sample from target distributions by learning a stochastic differential equation (SDEs) transport map that bridges between the target and a simple distribution. Specifically, for the problem of sampling from $q(x)$, consider the It\^o SDE,
\begin{equation}
    d\mathbf{x}_s = \mu_s \mathbf{x}_s\, ds + \varsigma_s  \,dW_s, \label{eq:lin_sde}
\end{equation}
describing $\mathbf{x}$ over time, $s \in [0, S]$, where $\mu_s$ and $\varsigma_s$ are the scalar coefficients, $W$ is the Wiener process, and $\mathbf{x}_0 \sim q(x)$. With the right coefficients, the stationary distribution can be  $p_S(x_S) \approx \mathcal{N}(\mathbf{0}, \cdot)$. Sampling is performed by finding a reverse process with the same marginals as the forward process. A reverse process, $\bar{\mathbf{x}}_s$, can be expressed in terms of the score function~\cite{anderson1982reverse}, $\nabla_{\mathbf{x}_s} \log p_s(\mathbf{x}_s)$,
\begin{equation}
    d \bar{\mathbf{x}}_s = \left[\mu_s \bar{\mathbf{x}}_s - \varsigma_s^2 \nabla \log p_s(\bar{\mathbf{x}}_s)\right] \,ds + \varsigma_s  \,d\bar W_s,
\end{equation}
where we write the score as $\nabla \log p_s(\mathbf{x}_s)$ for convenience. Other popular discrete time processes such as DDPM can be equivalently described as reverse SDEs~\cite{song2021scorebased}. In addition, SDEs also admit deterministic Probability Flow ODEs (PF-ODEs) with the same marginals expressible with the score.

Generally, a score neural network, $\psi_\theta(\mathbf{x}, s)$ with parameters $\theta$, is trained to approximate  $\nabla \log p_s(\mathbf{x}_s)$ by sampling from the dataset, $\mathcal{D}$, running the forward noising process, and minimizing a variational lower bound or score matching loss, $\mathcal{L}(\cdot)$,
\begin{equation}
    \theta^* = \arg \min_\theta \mathop{\mathbb{E}}_{\mathbf{x}_0 \sim \mathcal{D}} ~\mathop{\mathbb{E}}_{\substack{s \in [0, S] \\x_s\sim p(x_s|x_0)}} \left[ \mathcal{L}(\mathbf{x}_s, \psi_\theta(\mathbf{x}_s,  s))\right]. \label{eq:loss}
\end{equation}

\subsection{Closed-Form Diffusion Models}
However, for a finite dataset, $\mathcal{D} = \{\hat{\mathbf{x}}_i\}^N_0$, the score function is analytically tractable to calculate. For this finite dataset, the initial empirical distribution is a mixture of deltas, $q(\mathbf{x}) = \frac{1}{N}\sum_i \delta(\mathbf{x} - \hat{\mathbf{x}}_i)$.

For a linear noising process, the push forward distribution conditioned on a given starting data point, $p_s(\mathbf{x}_s|\mathbf{x}_0)$, is given by a scaling and a Gaussian noising of the initial delta, both of which depend on the diffusion time, $s$. 
Specifically, for the mixture of deltas corrupted with a forward signal gain, $G_s$, and noising variance $\sigma_s^2$ such that the forward distribution is $p_s(\mathbf{x}_s\mid \mathbf{x}_0) \sim \mathcal{N}(G_s\mathbf{x}_0, \sigma_s^2)$, the time-dependent score for the marginalized mixture of Gaussians can be shown to be
\begin{equation}
    \nabla \log p_s(\mathbf{x})  = \frac{1}{\sigma_s^2}\left(k(\mathbf{x}) - \mathbf{x}\right), \text{ where} \ \  k(\mathbf{x}) = \sum_{i=1}^N\text{softmax}\left(\frac{-1}{2 \sigma_s^2} ||\mathbf{x} - G_s\hat{\mathbf{x}}_i||^2\right)_i G_s \hat{\mathbf{x}}_i, \label{eq:CF-Score}
\end{equation}

and the $\text{softmax}(\cdot)$ is taken over the $i$ dataset samples. This closed-form score can be used to run the diffusion process without training a neural network, as long as the dataset is available at inference.  

However, it is known that with this closed-form score, diffusion models show memorization to existing points in the dataset \cite{scarvelisclosed, Biroli2024}. As has been demonstrated in literature, neural diffusion models actually show generalization to novel samples due to the approximation error of the score networks due to a variety of effects such as equivariance and locality constraints in the approximator~\cite{kamb2024analytic}, the optimality gap in the loss function due to underfitting~\cite{song2025underfitting}, and the inductive biases of neural networks~\cite{scarvelisclosed}. 

Based on the fact that neural networks show an inductive bias towards smoothing, \citet{scarvelisclosed} showed that diffusion models could be run without neural networks and show generalization by artificially introducing errors to the process through Monte Carlo Gaussian smoothing, which provides generalization through an inductive bias towards sampling barycenters of training points. 

\section{Methodology}
We aim to learn a policy for a dynamical system with $D_a$-dimensional inputs and $D_o$-dimensional outputs. Similar to previous works in IL, we employ chunking where parametrize the action space as $N_a$-sized steps chunks of system inputs, $\mathcal{A} \subseteq \mathbb{R}^{N_a\times D_a}$, and observation space as $N_o$-sized chunks, $\mathcal{O} \subseteq \mathbb{R}^{N_o\times D_o}$. Observations may be state measurements, or a vector history of observations for a partially observable system. We aim to do IL in a probabilistic inference framework where we sample actions, $a \in \mathcal{A}$, based on current observations, $o \in \mathcal{O}$ by sampling from a conditional distribution, $p(a \mid o)$, based on a collected dataset of expert demonstrations, $\mathcal{D} = \{(\hat a_i, \hat o_i)\}_{i=0}^N$. 

\subsection{Closed-Form Diffusion Policy}
We propose using the smoothed closed-form score to sample actions based on the expert demonstration dataset using a diffusion process. However, the closed-form diffusion models are generally limited to unconditional generation. Policies require conditional generation to sample from $p(a\mid o)$. When $o$ is discrete, it is simple to run a closed-form diffusion model by partitioning our dataset and evaluating the score for each possible value of $o$. However, for a finite dataset of continuous observations, we must interpolate in between datasets by smoothing.

\paragraph{Conditional Kernel Estimation.}
To generate samples from $p(a \mid o)$, we need the conditional score $\nabla_{a_s} \log p_s(a_s \mid o)$. We derive this directly from a non-parametric model of the joint distribution, $p(a, o)$. 
Given a dataset of observation--action pairs $\mathcal{D} = \{(\hat o_i, \hat a_i)\}_{i=1}^N$, the empirical distribution is given by a mixture of deltas, $p_\mathcal{D}(a, o) = \frac{1}{N} \sum_{i=1}^N \delta (a - \hat a_i)\delta (o - \hat o_i)$. 

Instead, we model the joint distribution by keeping the action dimension as a mixture of deltas, mapping the observations to latent features $z = \phi(o)$ through a feature map $\phi: \mathcal{O} \to \mathcal{H}$ and smoothing with a Gaussian kernel, $K_h$, with bandwidth $h$, 
\begin{equation}
    p(a, o) = \frac{1}{N}\sum_{i=1}^N \delta(a - \hat a_i) K_h(\phi(o) - \phi(\hat o_i)) = \frac{1}{N}\sum_{i=1}^N \delta(a - \hat a_i) K_h(z - \hat z_i).
    \label{eq:joint_kde}
\end{equation}
Throughout the rest of this work, we work directly with $z$ rather than the raw observation $o$.
\paragraph{Conditional Smoothed Closed-Form Score.} Based on Equation \eqref{eq:joint_kde}, we can find the Monte Carlo-smoothed score function to take the form of Equation \eqref{eq:CF-Score} except with a modified $k(\cdot)$
\begin{equation}
    k(a_s, z) = \frac{1}{M}\sum_{j=1}^M\left[\sum_{i=1}^N
        \operatorname{softmax}\left(
            -\frac{\|a_s + \epsilon_j  - G_s \hat a_i\|^2}{2\sigma_s^2}
            -\frac{\|z - \hat z_i\|^2}{2h^2}
        \right)_{i}
        G_s \hat a_i\right].
    \label{eq:smoothed_cond_kernel}
\end{equation}
where $\epsilon_i \sim \mathcal{N}(0, \tau^2I)$ are i.i.d perturbations. Appendix \ref{app:closed_form_score} provides a derivation for this.

\paragraph{Mahalanobis Distance Kernel.} In general, $\phi(\cdot)$ may be learned (e.g. a neural network embedding) or specified analytically. For this work, we specialize our kernel to a local Mahalanobis distance kernel to account for non-uniform units and data geometry. We use a kNN query to fetch only the $k_{nn}$ nearest observations (in a Euclidean metric) before computing the score to calculate the local covariance. This controls the locality of the Mahalanobis kernel, on top of speeding up computation. Appendix \ref{app:whitening} elaborates the universality of the kernel approach and our choice of kernel.

\paragraph{Diffusion Process.} Equation \eqref{eq:smoothed_cond_kernel} is compatible with any linear forward noising process. Following \cite{scarvelisclosed}, we use the process that linearly interpolates data and noise~\cite{liu2023flow} between $s\in[0,1]$, giving forward gain and noise schedule $G_s = 1-s$ and $\sigma_s = s$. We run a Euler discretized version of the PF-ODE (See Appendix \ref{app:rectified_flow_ode} for derivation) in reverse,
\begin{equation}
    \frac{\mathrm{d} a_s}{\mathrm{d} s}
      =  \frac{-1}{1-s}\left[a_s + s\nabla_{a}\log p_s(a_s \mid z)\right],
    \label{eq:flow_ode}
\end{equation}
initialised at $a_1 \sim \mathcal{N}(0, I)$. We discretize time and map continuous time to indices, $[0, 1] \mapsto [0..S]$.
\begin{algorithm}[t]
    \label{algo:cfdp}
    \caption{Closed-Form Diffusion Policy}
    
    \KwIn{Dataset $\mathcal{D} = \{\hat a_i, \hat o_i\}_0^N$, CFDP parameters ($h, \tau, k_{nn}$, $N_a$, $N_o$), Feature map $\phi$}
    Construct k-NN tree ($\mathcal{T}$) from dataset ($\mathcal{D}$)

    \While{$t \gets t + H$}{
        Sample $a_S \sim \mathcal{N}(0, I)$ \\
        $o \gets $ Sense and chunk observation history from robot \\
        Map observation $z \gets \phi(o)$ \\
        Query $\underline{\mathcal{D}} = \{\hat a_i, \hat z_i\}_{0}^{k_{nn}}$ from $\mathcal{T}$\\ 
        \For{$s \gets$ $S..1$ steps}{
            Compute $\nabla \log p_s(a_s \mid z)$ from $\underline{\mathcal{D}}$ according to \eqref{eq:smoothed_cond_kernel} \\
            $a_{s-1} \gets $ according to discretized diffusion process \eqref{eq:flow_ode}
        }
        $u_{t:t+H} \gets a_0$ \\
        Execute actions on robot for $H$ timesteps
    }
\end{algorithm}
This results in the Closed-Form Diffusion Policy algorithm as described in Algorithm \ref{algo:cfdp}. We provide further intuition through 1D demonstrations of CFDPs in Appendix \ref{app:simplified_demo}.

\paragraph{Hyperparameters} CFDPs allows control over amount of generalization over observations by controlling the bandwidth $h$, with small values of $h$ preferring actions with similar observations. With our kernel, $h$ takes the units of Mahalanobis distance. Similarly, $\tau$ controls generalization in the actions, with small values showing memorization and high values causing oversmoothing. Note that $\sigma_s$ is determined by the diffusion noise schedule.
\begin{wrapfigure}[14]{r}{0.45\textwidth}
\includegraphics[width=\linewidth]{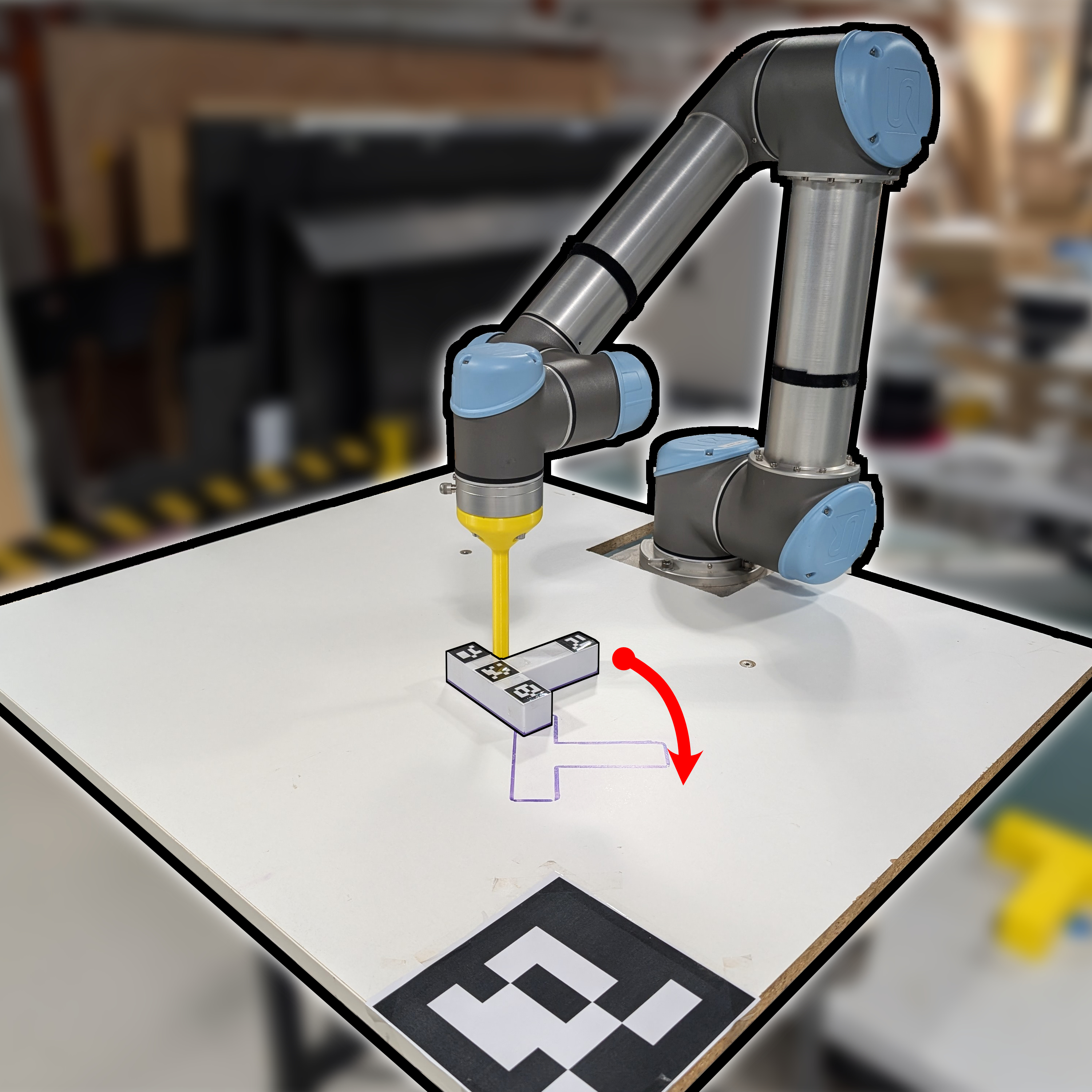}
\vspace{-3ex}\caption{PushT task on UR5e robot}
\label{fig:ur5e_hardware}
\end{wrapfigure}
\section{Hardware Experiments} 
We validate CFDPs on the state-based PushT hardware environment~\cite{chi_diffusion_2023}, where a UR5e robot pushes a \PushTsymbol-shaped block to a target pose (Figure~\ref{fig:ur5e_hardware}). The policy is trained exclusively on \emph{simulated} 2D data; pose estimates are obtained via fiducial markers.
 
 \paragraph{Task Performance.} 
 Despite being provided only simulation data, we find the CFDP policy successfully returns the block to the target pose 72\% of the time out of 25 episodes. The hardware environment had noisy pose estimation and visibly different dynamics due to friction variation. As the hardware environment did not have walls like simulation, the failures include episodes where the policy relied on reaction forces from the wall for manipulation.
 
\paragraph{Computational Performance.}
\begin{wraptable}[9]{r}{0.45\textwidth}
\begin{tabular}{@{}cccc@{}}
\toprule
\multicolumn{1}{l}{} & \multicolumn{1}{l}{\textbf{CPU}} & \multicolumn{1}{l}{\textbf{GPU}} & \multicolumn{1}{l}{\textbf{Pre-inference}} \\ \midrule
CFDP                 & \textbf{0.083 s}                 & \textbf{0.034 s}               & \textbf{0.501 s}                                             \\
NDP     & 0.637 s                          & 0.078 s                         & 15600 s                                                     \\ \bottomrule 
\end{tabular}
\vspace{-0.5ex}
\caption{Inference time on CPU and GPU, and pre-inference time including data loading, pre-processing, and (for NDP) model training.}
\label{tab:computational_results}
\end{wraptable}
Table \ref{tab:computational_results} shows a performance comparison of a CFDP ($k_{nn} = 2000$) and an NDP trained on the PushT dataset, both with $S=100$ diffusion steps. Appendix \ref{app:neural_policy_training} provides details on NDP training. We compare the pre-inference time (including training time), and inference time on both an RTX 3070 GPU and a mobile Intel Ultra 5 135U CPU (as a proxy for common embedded platforms). We find CFDP is \textbf{7x faster on CPU} and \textbf{2x faster on GPU} than the NDP for inference, on top of only requiring 0.5 seconds of pre-processing instead of the NDP's 4\,h\,20\,m of training on the RTX 3070 GPU. The NDP inference rate of 1.5\,Hz on CPU is too slow and causes jerky policy execution while the CFDP smoothly executes the task.

\section{Benchmark Experiments}
\begin{table}[t]
\centering
\includegraphics[width=1.0\linewidth]{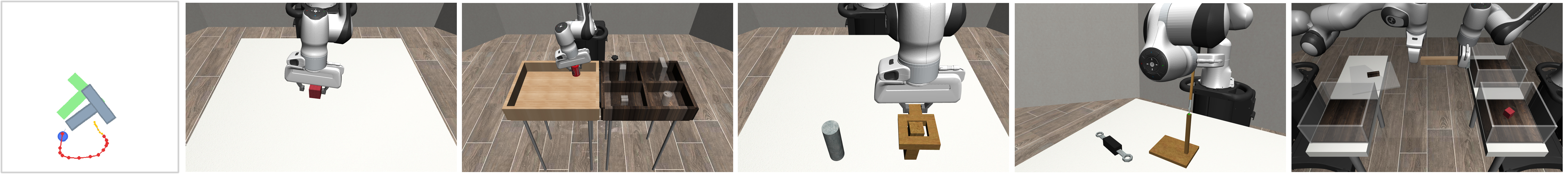}
\begin{tabular}{@{}lc|c|ccccc@{}}
\toprule
\textbf{Policy} &
  \multicolumn{1}{l|}{\textbf{Training?}} &
  \multicolumn{1}{l|}{\textbf{PushT}} &
  \multicolumn{1}{l}{\textbf{Lift}} &
  \multicolumn{1}{l}{\textbf{Can}} &
  \multicolumn{1}{l}{\textbf{Square}} &
  \multicolumn{1}{l}{\textbf{Transport}} &
  \multicolumn{1}{l}{\textbf{ToolHang}} \\ \midrule
CFDP (Ours)    &  \xmark        & 0.80 & 0.98 & 0.29 & 0.80 & 0.59 &  0.20    \\
1-NN Policy    &  \xmark       & 0.63 & 0.00 & 0.02 & 0.14 &  0.01    &  0.07    \\ \midrule
NDP-C~\cite{chi_diffusion_2023} & \cmark & 0.91 & 0.98 & 0.96 & 0.93 & 0.82 & 0.30 \\
NDP-T~\cite{chi_diffusion_2023} &\cmark  & 0.79 & 1.00 & 1.00 & 0.89 & 0.84 & 0.87 \\
LSTM-GMM~\cite{robomimic2021}   & \cmark                    & 0.61 & 0.96 & 0.91 & 0.73 & 0.47 & 0.31  \\
IBC~\cite{florence2021implicit} & \cmark                           & 0.84 & 0.41 & 0.00 & 0.00 & 0.00 & 0.00 \\
BET~\cite{shafiullah2022behavior}  & \cmark                          & 0.70 & 0.96 & 0.89 & 0.52 & 0.14 & 0.20 \\ \bottomrule
\end{tabular} 
\vspace{0.3em}
\caption{Results on state-based benchmarks. We compare against 1-NN policy (Appendix \ref{app:1nn_policy}), and baselines as reported by \cite{chi_diffusion_2023} for CNN and Transformer Diffusion Policies (NDP-C \& NDP-T), LSTM-GMM~\cite{robomimic2021}, Implicit BC (IBC)~\cite{florence2021implicit}, and Behavior Transformer (BET)~\cite{shafiullah2022behavior}.}
\label{tab:low-dim-results}
\end{table}
Table \ref{tab:low-dim-results} shows our results on state-based imitation benchmarks against neural baselines from \cite{chi_diffusion_2023}. We perform competitively against neural baselines on most tasks despite being training-free. We also compare to a training-free policy (1-NN Policy, Appendix \ref{app:1nn_policy}) that reproduces local data density-aware nearest actions. Appendix \ref{app:experimental_details} expands on the evaluation, experimental details and hyperparameter choices. Appendix \ref{sec:effect_of_hyperparams} provides studies of hyperparameter sensitivity, showing robustness to large variations in hyperparameter choices.

\paragraph{Benchmark Tasks.} We evaluate CFDPs on the 2D PushT environment~\cite{chi_diffusion_2023}, this time in simulation. Success is measured in area of the target pose covered as a percentage. Appendix \ref{app:pusht_rollouts} shows example rollouts for this environment. We also evaluate performance on state-based Robomimic benchmark tasks~\cite{robomimic2021} where the policy has access to end-effector and objects poses. Similar to \cite{chi_diffusion_2023}, we use absolute action spaces. The success metric is binary task successes as a percentage.

\paragraph{CFDPs are competitive with neural baselines.} 
Despite requiring no training, CFDPs are competitive with neural baselines on most tasks. 
Although the inductive bias of the closed-form diffusion models is simple, weighted averages of robot actions are often reasonable new actions, unlike the image generation case where this bias on its own is insufficient and requires methods such as latent spaces~\cite{scarvelisclosed}. 
Consequently, the performance bottleneck for CFDPs is modeling the conditional geometry of how close a given observation is to other demonstrations in the dataset.
\paragraph{CFDPs are strong surrogates for neural diffusion policies.} Having the same underlying process, CFDPs share many properties of NDPs. The performance of NDPs and CFDPs are generally correlated. When NDP-C struggles in ToolHang, CFDP does too. On the other hand, neural policies like IBC and BET have catastrophic degradation in various tasks where NDPs and CFDPs don't\footnote{This is not a guarantee however, as CFDPs struggled more on the Can task -- we believe this may be because the initial distribution of positions of the can is much larger.}. Additionally, CFDPs share many subtle properties to NDPs, e.g. in sparse regions of the observation space, CFDPs show memorization like NDPs~\cite{he2025demystifying} (since the Eq \eqref{eq:cond_score} is dominated by the nearest data point) but they smooth when there's multiple similar observations. We found that both CFDPs and NDPs shared a similar looping failure mode on the PushT task when the ~\PushTsymbol~ was on the top left quarter of the screen, showing that CFDPs can act as a surrogate for finding failures with NDPs.

We suggest that CFDPs can serve as diagnostic and mechanistic surrogate for studying the behavior of NDPs by identifying policy data coverage, identifying difficult tasks, or providing an idealized score-based policy to show the effect of data on a policy.
\section{Inference-time Neural Policy Editing}\label{sec:policy_editing}
We show that not only are closed-form scores able to produce training-free policies via CFDP, but act as a composable primitive that can edit the behavior of pre-trained NDP at inference-time using only data. In particular, we show that we can guide NDPs towards specific actions or extend their coverage by augmenting them with novel demonstrations.
\begin{figure}[t]
    \centering
    \includegraphics[width=\linewidth]{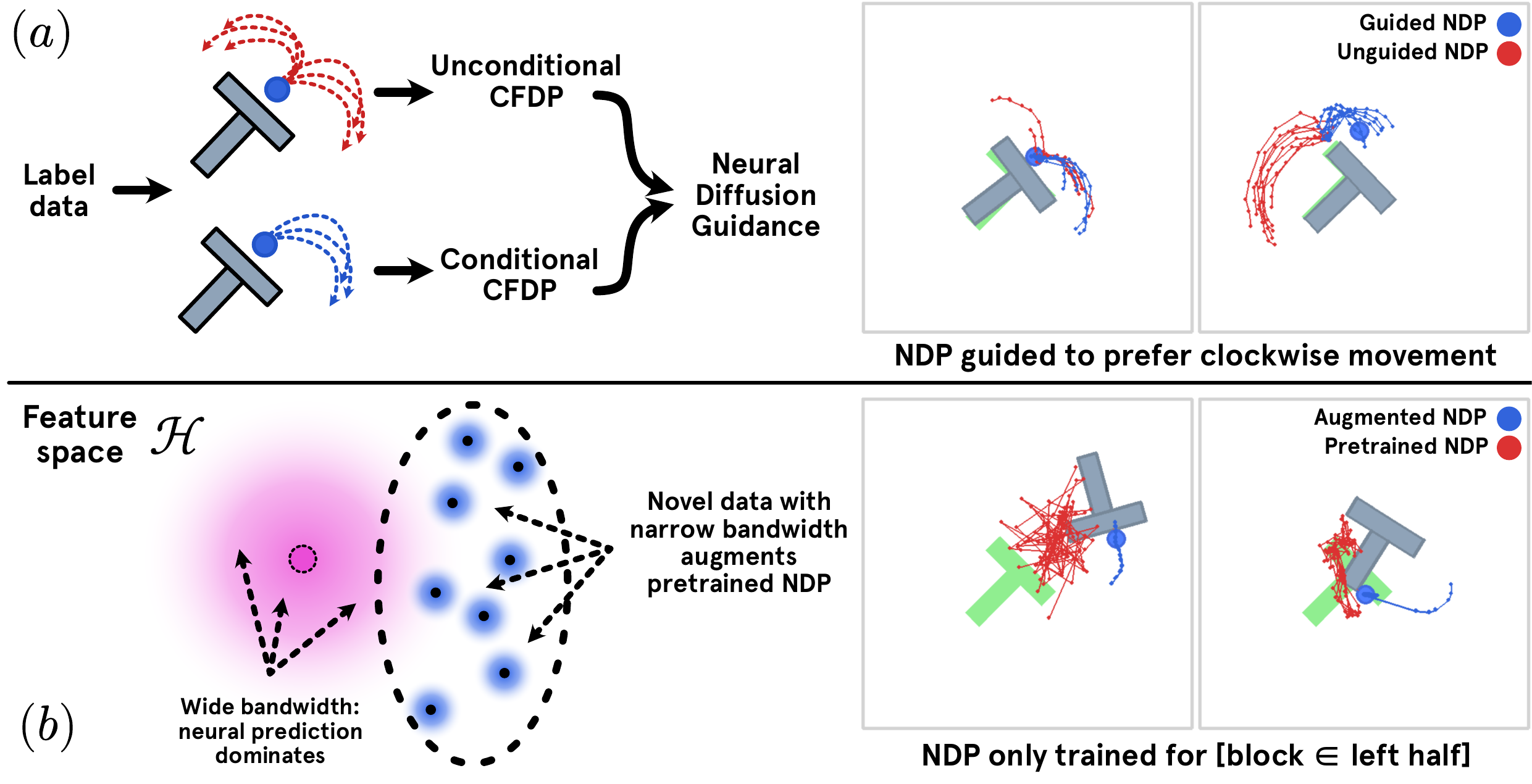}
    \caption{(a) A conditional-unconditional CFDP pair can guide neural diffusion policies to steer the action distribution based on demonstrations (b) CFDP can augment NDP for out-of-distribution observations without hand-crafted policy switching logic}
    \label{fig:neural_editing}
\end{figure}
\subsection{Data-driven Policy Guidance} \label{sec:policy_guidance}
It may be desirable to steer the generated actions from a pre-trained NDP towards specific actions. CFDPs can guide pre-trained NDPs without specifying rewards or modifying training by providing conditional scores for Classifier-Free Guidance (CFG). To guide an NDP, we can collect and label demonstration trajectories that match a given desired condition, $c$, and create a guided score,
\begin{equation}
    \nabla \log p_s(a_s \mid z, c) := \nabla \log p_s^{NN}(a_s \mid z) + w[\nabla \log p_s^{CF}(a_s\mid z, c) - \nabla \log p_s^{CF}(a_s \mid z)],\label{eq:cfdp_cfg}
\end{equation}
where we denote $NN$ and $CF$ for the terms associated with the neural and closed-form scores respectively, and $w$ is a guidance strength. Appendix \ref{app:policy_edit_guidance} provides a derivation. The conditional closed-form score, $\nabla \log p_s^{CF}(a \mid z, c)$, is constructed from only the $c$-labeled demonstrations.

We show experiments where we manually partition the PushT dataset to filter out chunks of demonstrations where the agent cursor moves counter-clockwise after being near the head of the ~\PushTsymbol~ block (Figure \ref{fig:neural_editing}).
Appendix \ref{app:neural_editing} provides further details on the data labeling.
We use the partitioned dataset to guide the NDP to move clockwise when the cursor is near the head. Table \ref{tab:cfg-results} shows our results, showing that the CFDP-guided NDP provides higher performance than using a CFDP trained on the filtered dataset (Conditional CFDP) and nearly as good as the pre-trained NDP that does not match the condition. Figure \ref{fig:neural_editing}(a) shows how the guidance steers actions towards clockwise movement.

\subsection{Novel Demonstration Augmentation} 

Once trained, one may want to edit an NDP to handle new out-of-distribution observations it was not trained on by collecting more data. Replacing the NDP with a CFDP constructed from the expanded dataset sacrifices superior performance, or may be infeasible if the entire dataset is not retained. We show that new demonstrations can instead be added to an NDP at inference time, \emph{without} finetuning. 

We add the contribution of the NDP as a virtual data point in the score computation. At each diffusion step, we find an action prediction $\mathbb{E}[a_0|a_s]$ by the model via Tweedie's formula. We place the neural data point in observation space as being centred at its empirical mean observation. This results in a new dataset,  $\mathcal{D}_{s}^{\text{aug}} = \mathcal{D} \cup \{(\mathbb{E}[a_0|a_s],\, \mu_o^{NN})\}$. Lastly, since the NDP has seen a variety of observations, its virtual data point must implicitly capture a wide range of observations--we represent this by calculating the observation kernel using a scaled empirical covariance of the whole neural training dataset, $c\Sigma^{NN}$, as the distance metric. We extend our conditional score to account for heterogeneous and non-isotropic Gaussian kernels, and this results in an extension of Eq. \eqref{eq:CF-Score} as 
\begin{equation}
    k(a_s, o) = \frac{1}{M}\sum_{j=1}^M\left[\sum_{i=1}^{N+1}
        \operatorname{softmax}\left(
            -\frac{\|a_s + \epsilon_j  - G_s \hat a_i\|^2}{2\sigma_s^2}
            + \log w_i^{\mathrm{obs}}
        \right)_{i}
        G_s \hat a_i\right]
    \label{eq:augmented_kernel}
\end{equation}
where
\begin{equation}
    \log w_i^{\mathrm{obs}} = -\tfrac{1}{2}(o-\mu_i)^{\top} H_i^{-1}(o-\mu_i) - \tfrac{1}{2}\log\det H_i,
\end{equation}
is an expanded form of the observation distance term from before that depends on the the type of data point. Here $H_i$ is an matrix that specializes to the scaled empirical covariance for the neural data point and the local dataset covariance for the novel demonstration points. Appendix \ref{app:augmentation} provides further details on the derivation and implementation. We note that with just the neural prediction, Equation \eqref{eq:augmented_kernel} collapses into just the neural prediction, and with just the closed-form dataset is equivalent to our local Mahalanobis kernel CFDP, naturally capturing both extremes of the hybrid policy.

We perform experiments on the PushT environment where we manually partition the demonstration dataset based on whether the center of the ~\PushTsymbol~ block is on the left or the right side of the screen. An NDP is trained on the left side dataset, and a CFDP is trained on the rest. We then combine the two policies into a single augmented policy that can incorporate both of their data. Table \ref{tab:demonstration-aug-results} shows that our Augmented NDP performs significantly better than the NDP or the CFDP that it is composed of. Figure \ref{fig:neural_editing} (b) shows how with novel demonstration augmentation, the closed-form score takes over when the pre-trained NDP is out-of-distribution and producing poor actions. 

\begin{table}[t]
\centering

\begin{minipage}[t]{0.48\linewidth}
\centering
\captionof{table}{Data-driven Policy Guidance results}
\label{tab:cfg-results}
\vspace{0.5ex}

\begin{tabular}{@{}lll@{}}
\toprule
\textbf{Policy} & \textbf{Match Cond.} & \textbf{Success} \\ 
\midrule
\textbf{Pre-trained NDP}      & \xmark & \textbf{0.90} \\
\textbf{Conditional CFDP} & \cmark & 0.79 \\
\textbf{CFDP-Guided NDP}  & \cmark & 0.88 \\ 
\bottomrule
\end{tabular}
\end{minipage}
\hfill
\begin{minipage}[t]{0.48\linewidth}
\centering
\captionof{table}{Demonstration Augmentation results}
\label{tab:demonstration-aug-results}
\vspace{0.5ex}

\begin{tabular}{@{}lll@{}}
\toprule
\textbf{Policy}        & \textbf{Dataset} & \textbf{Performance} \\ 
\midrule
\textbf{Pre-trained NDP}           & Left             & 0.36 \\
\textbf{Novel data CFDP}          & Right            & 0.33 \\
\textbf{Augmented NDP} & Both             & \textbf{0.58} \\ 
\bottomrule
\end{tabular}
\end{minipage}

\end{table}

\section{Conclusion}
We introduced Closed-Form Diffusion Policies, a class of diffusion-based policies for imitation learning with instant deployment and real-time inference on constrained hardware. Our benchmarks show that despite being training free, CFDPs are competitive with neural baselines, representing a favorable tradeoff between training time and performance. We show closed-form scores are a composable primitive for practical inference-time policy editing without retraining. We suggest using CFDPs as a modular framework for rapid in-situ policy generation during data collection, as a mechanistic surrogate for characterizing diffusion policy behavior without training, and using them for inference-time editing to steer generated actions and patching in new behavior. 

We plan to extend this work to visuomotor policies using pretrained representations and lightweight learning methods. 
The inference speed of CFDPs also lends itself towards online IL methods such as DAgger~\cite{gordon2011_dagger} for use in collecting data for neural policies in the same deployment. 
Work regarding the implicit biases of neural diffusion models and how to replicate them~\cite{kamb2024analytic} are ongoing and could provide a direction for improved generalization.
\newpage
\section{Limitations}
\paragraph{Joint Modeling} As we model our observation-action joint distribution by mapping to a feature space and smoothing with kernels, we adopt the limitation of kernel methods and similar non-parametric estimators. In particular, curse of dimensionality for high dimensional modalities like images makes all points become nearly equidistant from one another. This can be addressed by incorporating feature maps that map to dense low dimensional representations but requires techniques to find representations that are useful for the task without additional training. We aim to extend this work with pre-trained embeddings and task-agnostic feature maps for visuomotor policies in future work. 

\paragraph{Failure Modes} The most common failure mode that we observe in CFDPs is the policy entering a loop due to being in a sparsely sampled part of the observation space, resulting in an action where the policy recommends the similar chunk even after a chunk is finished executing. We \emph{also} notice this behavior in NDPs, which makes CFDPs an excellent candidate for diagnosing sparse neighborhoods of data for NDPs. However we notice this behavior is more common in CFDPs. Higher observation kernel bandwidths can reduce this, however that can negatively impact performance by oversmoothing, causing lack of precision. We find CFDPs generally benefit from longer action chunk sizes as it ensures the state at the end of the chunk is more distinct from previous observations, reducing looping. 

\paragraph{Inflexible Inductive Biases} While our performance shows entirely training-free methods are enough to outperform many neural baselines, the inductive bias of Monte Carlo score smoothing is to sample in the convex hull of training points. This is a much more appropriate bias in control problems, where barycenters of control actions are usually reasonable actions, than domains like image generation. However, modern neural networks are still stronger learners with diverse inductive biases. This could potentially be worked on by recent works probing the generalization biases of neural networks in diffusion models~\cite{kamb2024analytic} or learning latent action representations where interpolation is more meaningful~\cite{scarvelisclosed}.

\paragraph{Dataset Size} Our method requires the ability to store and query the training dataset. This is not generally very limiting, as usually imitation learning data is collected on a single robot anyway which can then be instantly deployed onto a CFDPs. However, this means our method can not be easily applied to internet-scale data as is used in foundation models. CFDPs can still be used for editing the behavior of a pre-trained diffusion-based foundation policy as shown in Section \ref{sec:policy_editing}.

\paragraph{Manual Tuning} Compared to NDPs, as with many kernel methods, we find CFDPs require more task-specific tuning, especially the bandwidth, $h$. We don't find this prohibitive as (1) there is no training time, which saves significant time for tuning (2) there are only a few policy parameters, and (3) their roles are very interpretable, making the tuning process simple. 

\paragraph{Neural Data Augmentation} Using a Gaussian kernel in the observation feature space to distill the training distribution of an NDP is a simplifying assumption that doesn't capture the complexity of the true training data distribution which is in general non-Gaussian and multi-modal. However, accurately modeling this is a difficult problem of out-of-distribution detection that is an active research area and outside the scope of our work. We leave more complex observation distribution modeling for improved augmentation performance to future research.

\acknowledgments{This research was funded by the Australian Research Council through the ARC Research Hub in Intelligent Robotic Systems for Real-Time Asset Management (IH210100030). All authors are with the Australian Robotic Inspection and Asset Management (ARIAM) Hub, the Australian Centre for Robotics, and the School of Aerospace, Mechanical and Mechatronic Engineering, University of Sydney.}
\newpage
\bibliography{references}  

@article{he2025demystifying,
  title={Demystifying Diffusion Policies: Action Memorization and Simple Lookup Table Alternatives},
  author={He, Chengyang and Liu, Xu and Camps, Gadiel Sznaier and Sartoretti, Guillaume and Schwager, Mac},
  journal={arXiv preprint arXiv:2505.05787},
  year={2025}
}

@article{anderson1982reverse,
  title={Reverse-time diffusion equation models},
  author={Anderson, Brian D O},
  journal={Stochastic Processes and their Applications},
  volume={12},
  number={3},
  pages={313--326},
  year={1982},
  publisher={Elsevier}
}

@inproceedings{
  song2021scorebased,
  title={Score-Based Generative Modeling through Stochastic Differential Equations},
  author={Yang Song and Jascha Sohl-Dickstein and Diederik P Kingma and Abhishek Kumar and Stefano Ermon and Ben Poole},
  booktitle={International Conference on Learning Representations},
  year={2021},
  url={https://openreview.net/forum?id=PxTIG12RRHS}
}

@inproceedings{robomimic2021,
  title={What Matters in Learning from Offline Human Demonstrations for Robot Manipulation},
  author={Ajay Mandlekar and Danfei Xu and Josiah Wong and Soroush Nasiriany and Chen Wang and Rohun Kulkarni and Li Fei-Fei and Silvio Savarese and Yuke Zhu and Roberto Mart\'{i}n-Mart\'{i}n},
  booktitle={arXiv preprint arXiv:2108.03298},
  year={2021}
}

@inproceedings{
      shafiullah2022behavior,
      title={Behavior Transformers: Cloning $k$ modes with one stone},
      author={Nur Muhammad Mahi Shafiullah and Zichen Jeff Cui and Ariuntuya Altanzaya and Lerrel Pinto},
      booktitle={Thirty-Sixth Conference on Neural Information Processing Systems},
      year={2022},
      url={https://openreview.net/forum?id=agTr-vRQsa}
}

@article{florence2021implicit,
    title={Implicit Behavioral Cloning},
    author={Florence, Pete and Lynch, Corey and Zeng, Andy and Ramirez, Oscar and Wahid, Ayzaan and Downs, Laura and Wong, Adrian and Lee, Johnny and Mordatch, Igor and Tompson, Jonathan},
    journal={Conference on Robot Learning (CoRL)},
    year={2021}
}

@InProceedings{huang25_diffusion_learns_mpc,
  title = 	 {Toward Near-Globally Optimal Nonlinear Model Predictive Control via Diffusion Models},
  author =       {Huang, Tzu-Yuan and Lederer, Armin and Hoischen, Nicolas and Brudigam, Jan and Xiao, Xuehua and Sosnowski, Stefan and Hirche, Sandra},
  booktitle = 	 {Proceedings of the 7th Annual Learning for Dynamics \&amp; Control Conference},
  pages = 	 {777--790},
  year = 	 {2025},
  editor = 	 {Ozay, Necmiye and Balzano, Laura and Panagou, Dimitra and Abate, Alessandro},
  volume = 	 {283},
  series = 	 {Proceedings of Machine Learning Research},
  month = 	 {04--06 Jun},
  publisher =    {PMLR},
  url = 	 {https://proceedings.mlr.press/v283/huang25a.html}
}

@InProceedings{li25_diffusolve,
  title = 	 {DiffuSolve: Diffusion-based Solver for Non-convex Trajectory Optimization},
  author =       {Li, Anjian and Ding, Zihan and Dieng, Adji Bousso and Beeson, Ryne},
  booktitle = 	 {Proceedings of the 7th Annual Learning for Dynamics \&amp; Control Conference},
  pages = 	 {45--58},
  year = 	 {2025},
  editor = 	 {Ozay, Necmiye and Balzano, Laura and Panagou, Dimitra and Abate, Alessandro},
  volume = 	 {283},
  series = 	 {Proceedings of Machine Learning Research},
  month = 	 {04--06 Jun},
  publisher =    {PMLR},
  url = 	 {https://proceedings.mlr.press/v283/li25a.html}
}

@inproceedings{li24_constraintawarediffusion,
author = {Li, Anjian and Ding, Zihan and Dieng, Adji Bousso and Beeson, Ryne},
title = {Constraint-Aware Diffusion Models for Trajectory Optimization},
year = {2024},
isbn = {978-3-031-94894-7},
publisher = {Springer-Verlag},
address = {Berlin, Heidelberg},
url = {https://doi.org/10.1007/978-3-031-94895-4_32},
doi = {10.1007/978-3-031-94895-4_32},
booktitle = {Dynamic Data Driven Applications Systems: 5th International Conference, DDDAS/Infosymbiotics for Reliable AI 2024, New Brunswick, NJ, USA, November 6–8, 2024, Proceedings},
pages = {308–316},
numpages = {9},
location = {New Brunswick, NJ, USA}
}

@inproceedings{du2025dynaguidesteeringdiffusionpolices,
    title={DynaGuide: Steering Diffusion Policies with Active Dynamic Guidance},
    author={Maximilian Du and Shuran Song},
    booktitle={Proceedings of the 39th Conference on Neural Information Processing Systems (NeurIPS)},
    year={2025}
}

@inproceedings{liu2023flow,
  title={Flow Straight and Fast: Learning to Generate and Transfer Data with Rectified Flow},
  author={Liu, Xingchao and Gong, Chengyue and Liu, Qiang},
  booktitle={The Eleventh International Conference on Learning Representations (ICLR)},
  year={2023}
}

@misc{mishra2026_ebmbd,
      title={{EB-MBD}: Emerging-Barrier Model-Based Diffusion for Safe Trajectory Optimization in Highly Constrained Environments}, 
      author={Raghav Mishra and Ian R. Manchester},
      year={2026},
      eprint={2510.07700},
      archivePrefix={arXiv},
      primaryClass={cs.RO},
      url={https://arxiv.org/abs/2510.07700}, 
}

@article{
scarvelisclosed,
title={Closed-Form Diffusion Models},
author={Christopher Scarvelis and Haitz S{\'a}ez de Oc{\'a}riz Borde and Justin Solomon},
journal={Transactions on Machine Learning Research},
issn={2835-8856},
year={2025},
url={https://openreview.net/forum?id=JkMifr17wc},
note={}
}

@article{song2025underfitting,
  title={Selective Underfitting in Diffusion Models},
  author={Song, Kiwhan and Kim, Jaeyeon and Chen, Sitan and Du, Yilun and Kakade, Sham and Sitzmann, Vincent},
  journal={arXiv preprint arXiv:2510.01378},
  year={2025}
}

@article{kamb2024analytic,
  title={An analytic theory of creativity in convolutional diffusion models},
  author={Kamb, Mason and Ganguli, Surya},
  journal={arXiv preprint arXiv:2412.20292},
  year={2024}
}

@article{wang2022diffusionRL,
  title={Diffusion policies as an expressive policy class for offline reinforcement learning},
  author={Wang, Zhendong and Hunt, Jonathan J and Zhou, Mingyuan},
  journal={arXiv preprint arXiv:2208.06193},
  year={2022}
}

@inproceedings{zhang_action_theory,
  title={Action Chunking and Data Augmentation Yield Exponential Improvements in Behavior Cloning for Continuous Spaces},
  author={Zhang, Thomas TCK and Pfrommer, Daniel and Pan, Chaoyi and Matni, Nikolai and Simchowitz, Max},
  booktitle={The Fourteenth International Conference on Learning Representations},
  year={2025}
}

@INPROCEEDINGS{Zhao_ACT, 
    AUTHOR    = {Tony Z. Zhao AND Vikash Kumar AND Sergey Levine AND Chelsea Finn}, 
    TITLE     = {{Learning Fine-Grained Bimanual Manipulation with Low-Cost Hardware}}, 
    BOOKTITLE = {Proceedings of Robotics: Science and Systems}, 
    YEAR      = {2023}, 
    ADDRESS   = {Daegu, Republic of Korea}, 
    MONTH     = {July}, 
    DOI       = {10.15607/RSS.2023.XIX.016} 
}

@InProceedings{gordon2011_dagger,
  title = 	 {A Reduction of Imitation Learning and Structured Prediction to No-Regret Online Learning},
  author = 	 {Ross, Stephane and Gordon, Geoffrey and Bagnell, Drew},
  booktitle = 	 {Proceedings of the Fourteenth International Conference on Artificial Intelligence and Statistics},
  pages = 	 {627--635},
  year = 	 {2011},
  editor = 	 {Gordon, Geoffrey and Dunson, David and Dudík, Miroslav},
  volume = 	 {15},
  series = 	 {Proceedings of Machine Learning Research},
  address = 	 {Fort Lauderdale, FL, USA},
  month = 	 {11--13 Apr},
  publisher =    {PMLR},
  url = 	 {https://proceedings.mlr.press/v15/ross11a.html}
}

@article{cleandiffuser,
  title={Cleandiffuser: An easy-to-use modularized library for diffusion models in decision making},
  author={Dong, Zibin and Yuan, Yifu and Hao, Jianye and Ni, Fei and Ma, Yi and Li, Pengyi and Zheng, Yan},
  journal={Advances in Neural Information Processing Systems},
  volume={37},
  pages={86899--86926},
  year={2024}
}

@inproceedings{xue2025full,
  title={Full-order sampling-based mpc for torque-level locomotion control via diffusion-style annealing},
  author={Xue, Haoru and Pan, Chaoyi and Yi, Zeji and Qu, Guannan and Shi, Guanya},
  booktitle={2025 IEEE International Conference on Robotics and Automation (ICRA)},
  pages={4974--4981},
  year={2025},
  organization={IEEE}
}

@article{pan_model-based_2024,
	title = {Model-based diffusion for trajectory optimization},
	volume = {37},
	journal = {Advances in Neural Information Processing Systems},
	author = {Pan, Chaoyi and Yi, Zeji and Shi, Guanya and Qu, Guannan},
	year = {2024},
	pages = {57914--57943},
}

@inproceedings{janner_planning_2022,
	title = {Planning with {Diffusion} for {Flexible} {Behavior} {Synthesis}},
	booktitle = {International {Conference} on {Machine} {Learning}},
	author = {Janner, Michael and Du, Yilun and Tenenbaum, Joshua and Levine, Sergey},
	year = {2022},
}

@misc{carvalho_motion_2024,
	title = {Motion {Planning} {Diffusion}: {Learning} and {Planning} of {Robot} {Motions} with {Diffusion} {Models}},
	shorttitle = {Motion {Planning} {Diffusion}},
	url = {http://arxiv.org/abs/2308.01557},
	language = {en},
	urldate = {2024-08-19},
	publisher = {arXiv},
	author = {Carvalho, Joao and Le, An T. and Baierl, Mark and Koert, Dorothea and Peters, Jan},
	month = mar,
	year = {2024},
	note = {arXiv:2308.01557 [cs]},
}

@inproceedings{ross2011reduction,
  title={A reduction of imitation learning and structured prediction to no-regret online learning},
  author={Ross, St{\'e}phane and Gordon, Geoffrey and Bagnell, Drew},
  booktitle={Proceedings of the fourteenth international conference on artificial intelligence and statistics},
  pages={627--635},
  year={2011},
  organization={JMLR Workshop and Conference Proceedings}
}

@inproceedings{jin_stage-wise_2025,
  author = {Jin, Cheng and Shi, Qitan and Gu, Yuantao},
  title = {Stage-wise {Dynamics} of {Classifier}-{Free} {Guidance} in {Diffusion} {Models}},
  year = {2026},
  booktitle = {International Conference on Learning Representations},
  url = {https://openreview.net/forum?id=fP0s1TEow3},
  language = {en},
  urldate = {2026-03-27},
  month = oct,
}

@Article{Biroli2024,
author="Biroli, Giulio
and Bonnaire, Tony
and de Bortoli, Valentin
and M{\'e}zard, Marc",
title="Dynamical regimes of diffusion models",
journal="Nature Communications",
year="2024",
month="Nov",
day="17",
volume="15",
number="1",
pages="9957",
issn="2041-1723",
doi="10.1038/s41467-024-54281-3",
url="https://doi.org/10.1038/s41467-024-54281-3"
}

@inproceedings{chi_diffusion_2023,
    title = {Diffusion {Policy}: {Visuomotor} {Policy} {Learning} via {Action} {Diffusion}},
    booktitle = {Proceedings of {Robotics}: {Science} and {Systems} ({RSS})},
    author = {Chi, Cheng and Feng, Siyuan and Du, Yilun and Xu, Zhenjia and Cousineau, Eric and Burchfiel, Benjamin and Song, Shuran},
    year = {2023},
}

@article{ho_denoising_2020,
    title = {Denoising {Diffusion} {Probabilistic} {Models}},
    language = {en},
    journal = {Advances in Neural Information Processing Systems},
    author = {Ho, Jonathan and Jain, Ajay and Abbeel, Pieter},
    year = {2020},
}

@misc{sohl-dickstein_deep_2015,
    title = {Deep {Unsupervised} {Learning} using {Nonequilibrium} {Thermodynamics}},
    url = {http://arxiv.org/abs/1503.03585},
    doi = {10.48550/arXiv.1503.03585},
    language = {en},
    urldate = {2025-09-04},
    publisher = {arXiv},
    author = {Sohl-Dickstein, Jascha and Weiss, Eric A. and Maheswaranathan, Niru and Ganguli, Surya},
    month = nov,
    year = {2015},
    note = {arXiv:1503.03585 [cs]},
}

@article{nesterov_random_2017,
    title = {Random {Gradient}-{Free} {Minimization} of {Convex} {Functions}},
    volume = {17},
    issn = {1615-3383},
    url = {https://doi.org/10.1007/s10208-015-9296-2},
    doi = {10.1007/s10208-015-9296-2},
    language = {en},
    number = {2},
    urldate = {2024-12-21},
    journal = {Foundations of Computational Mathematics},
    author = {Nesterov, Yurii and Spokoiny, Vladimir},
    month = apr,
    year = {2017},
    pages = {527--566},
}

\appendix
\newpage
\section{Conditional Closed-Form Score}
\label{app:closed_form_score}
With the modelling assumption in Eq. \eqref{eq:joint_kde}, the marginal over actions of the joint distribution is $p(z) = \frac{1}{N}\sum_i K_h(z - z_i)$, so the action distribution conditioned on latent observation mapping is
\begin{equation}
    p(a \mid z) = \frac{\sum_{i=1}^N \delta(a - \hat a_i)K_h(z - z_i)}
                         {\sum_{j=1}^N K_h(z - z_j)}.
    \label{eq:cond_kde}
\end{equation}

\paragraph{Corrupted conditional distribution.}
We apply the diffusion forward process to the action dimension only. With signal gain $G$ and noise variance $\sigma^2$, the conditional noised distribution is
\begin{equation}
    p(a_s \mid a_0) = (2\pi\sigma^2)^{-D_a/2}\exp\left(-\frac{\|a_s - G a_0\|^2}{2\sigma^2}\right),
    \label{eq:fwd_gaussian}
\end{equation}
and the Gaussian observation kernel $K_h$ used in \eqref{eq:joint_kde} is
\begin{equation}
    K_h(z - \hat z_i) = (2\pi h^2)^{-D_o/2}\exp\left(-\frac{\|z - \hat z_i\|^2}{2h^2}\right).
    \label{eq:obs_kernel_gaussian}
\end{equation}
Applying \eqref{eq:fwd_gaussian} only on the action dimension of the empirical joint $p(a, z) = \frac{1}{N}\sum_i \delta(a-\hat a_i)K_h(z-\hat z_i)$ yields the corrupted joint
\begin{equation}
    p_s(a_s, z) = \frac{1}{N}\sum_{i=1}^N (2\pi\sigma^2)^{-D_a/2}\exp\left(-\frac{\|a_s - G \hat a_i\|^2}{2\sigma^2}\right)K_h(z - \hat z_i),
    \label{eq:corrupted_joint}
\end{equation}
and integrating $a_s$ out leaves the marginal $p(z) = \frac{1}{N}\sum_i K_h(z-\hat z_i)$ untouched, so the corrupted conditional is
\begin{equation}
    p_s(a_s \mid z) =
        \frac{\displaystyle\sum_{i=1}^N (2\pi\sigma^2)^{-D_a/2}\exp\left(-\tfrac{\|a_s - G \hat a_i\|^2}{2\sigma^2}\right)K_h(z - \hat z_i)}
             {\displaystyle\sum_{j=1}^N K_h(z - \hat z_j)}.
    \label{eq:corrupted_cond}
\end{equation}

\paragraph{Deriving the conditional score.}
The denominator of \eqref{eq:corrupted_cond} is independent of $a_s$. Taking the $\log$ and differentiating, we get 
\begin{equation}
    \nabla_{a_s}\log p_s(a_s \mid z) = \sum_i w_i \frac{G \hat a_i - a_s}{\sigma^2},
\end{equation}
where the weights are
\begin{equation}
    w_i = \frac{(2\pi\sigma^2)^{-D_a/2}\exp\left(-\|a_s - G \hat a_i\|^2/(2\sigma^2)\right)K_h(z - \hat z_i)}
                    {\sum_j (2\pi\sigma^2)^{-D_a/2}\exp\left(-\|a_s - G \hat a_j\|^2/(2\sigma^2)\right)K_h(z -\hat  z_j)}.
    \label{eq:weights_raw}
\end{equation}
Since $\sum_i w_i = 1$, splitting the action term into a weighted average minus $a_s$ gives the final form
\begin{equation}
    \nabla_{a_s}\log p_s(a_s \mid z)
    = \frac{1}{\sigma^2}\left(\sum_i w_iG \hat a_i - a_s\sum_i w_i\right)
    = \frac{1}{\sigma^2}\left(k(a_s, z) - a_s\right),
    \label{eq:score_weighted_form}
\end{equation}
with $k(a_s, z) \equiv \sum_i w_iG \hat a_i$, matching the form of the unconditional CFDM score in \eqref{eq:CF-Score}. Pulling all factors inside a single exponential,
\begin{equation}
    w_i = \frac{\exp\left({-\tfrac{D_a}{2}\log 2\pi\sigma^2 - \tfrac{D_o}{2}\log 2\pi h^2} -\tfrac{\|a_s - G \hat a_i\|^2}{2\sigma^2} -\tfrac{\|z - \hat z_i\|^2}{2h^2}\right)}
         {\sum_j \exp\left(-\tfrac{D_a}{2}\log 2\pi\sigma^2 - \tfrac{D_o}{2}\log 2\pi h^2 -\tfrac{\|a_s - G a_j\|^2}{2\sigma^2} -\tfrac{\|z - z_j\|^2}{2h^2}\right)}.
    \label{eq:weights_with_normalisers}
\end{equation}
The two log-normalisers $\tfrac{D_a}{2}\log 2\pi\sigma^2$ and $\tfrac{D_o}{2}\log 2\pi h^2$ are constant in the summation index, so they are common offset and can be removed from the $\operatorname{softmax}(\cdot)$, leaving
\begin{equation}
    w_i = \operatorname{softmax}\left(
            -\frac{\|a_s - G \hat a_i\|^2}{2\sigma^2}
            -\frac{\|z - \hat z_i\|^2}{2h^2}
        \right)_{i}.
    \label{eq:weights_softmax}
\end{equation}
Combining \eqref{eq:score_weighted_form} with \eqref{eq:weights_softmax} gives the closed-form conditional score
\begin{equation}
    \nabla_{a_s}\log p_s(a_s \mid z)
        = \frac{1}{\sigma^2}\left(k(a_s, z) - a_s\right),
    \label{eq:cond_score}
\end{equation}
where
\begin{equation}
    k(a_s, o) = \sum_{i=1}^N
        \operatorname{softmax}\left(
            -\frac{\|a_s - G \hat a_i\|^2}{2\sigma^2}
            -\frac{\|z - \hat z_i\|^2}{2h^2}
        \right)_{i}
        G \hat a_i.
    \label{eq:cond_kernel}
\end{equation}
This has the same form as the unconditional score in \eqref{eq:CF-Score}, but with weights that jointly account for proximity in the action space (via the diffusion kernel) \emph{and} proximity in the observation space (via the observation kernel $K_h$). Intuitively, each training action $\hat a_i$ contributes to the score in proportion to how well it explains both the current noisy action $a_s$ and the query observation. We note that this is equivalent to performing Nadaraya-Watson kernel regression to find the score conditioned on observation feature, $z$, for the at each diffusion step.

Lastly, along similar lines as \cite{scarvelisclosed}, we perform smoothing in the action dimension via Monte Carlo smoothing to provide generalization
\begin{equation}
    k(a_s, o) = \frac{1}{M}\sum_{j=1}^M\left[\sum_{i=1}^N
        \operatorname{softmax}\left(
            -\frac{\|a_s + \epsilon_j  - G \hat a_i\|^2}{2\sigma^2}
            -\frac{\|z - \hat z_i\|^2}{2h^2}
        \right)_{i}
        G \hat a_i\right].
\end{equation}
where $\epsilon_i$ are samples taken from $\mathcal{N}(0, \tau^2I)$ as smoothing perturbations. 

\section{Feature Mapping}\label{app:whitening} 
\paragraph{Universality} Equation \eqref{eq:joint_kde} models each action as having a kernel that determines how much that action should be generalized to another observation, $o'$. In general, this could be arbitrary and observation-dependent, which can be represented with a general positive semi-definite kernel $\kappa(o, o')$. 

We note that any continuous positive-definite kernel $\kappa(o, o')$ on compact $\mathcal{O}$ is realized as $K_h(z - z')$ for some $\phi$, since by Mercer's theorem every such kernel admits a feature map $\phi: \mathcal{O} \to \mathcal{H}$ with $\kappa(o,o') = \langle \phi(o), \phi(o')\rangle_\mathcal{H}$. As the Gaussian RBF is a universal kernel, the family ${K_h(\phi(o) - \phi(o'))}$ is dense in all continuous observation kernels on $\mathcal{O}$. Therefore, with some assumptions, CFDPs can theoretically capture arbitrarily complex generalizations of a given demonstration observation to previously unseen observations. However this relies on having an existing feature map that meaningfully helps generalization. Future work aims to focus on task-agnostic representation learning and using pre-trained vision embeddings in the feature map in order to provide training-free visuomotor imitation learning.

\paragraph{Mahalanobis Distance Kernel} Specifically, we specialize $\phi(\cdot)$ to the local Mahalanobis distance kernel. For each $o$, we take subsampled dataset $\underline{\mathcal{D}}$ with the $k_{nn}$ nearest neighbours, find the inverse Cholesky factor of the local sample covariance matrix $\Sigma = \frac{1}{k_{nn}-1}(O - \bar{O})^\top(O - \bar{O}) + \lambda I$, where $O \in \mathbb{R}^{k_{nn} \times D_o}$ is the data matrix that stacks the neighbor conditioning vectors, $\bar{O}$ is their mean, and $\lambda$ is a small diagonal regularizer that keeps $\Sigma$ well-conditioned if neighbors span a degenerate subspace. 

Taking the Cholesky factorization $\Sigma = LL^\top$ and inverting yields $L^{-1}$. With this, we define the feature map, $z = \phi(o) = L^{-1}o$. This transforms maps the observations, $\hat o_i$, into $\hat z_i$ which has a standard isotropic covariance of $I$. This means the bandwidth, $h$, can be interpreted to be in Mahalanobis distance units. This feature mapping is also equivalent to a locally varying anisotropic metric on the original observations as $||z - z'|| = ||L^{-1}o - L^{-1}o'|| = (o - o')^T L^{-T}L^{-1}(o - o') = ||o - o'||_{\Sigma^{-1}}$. 
\section{Flow Process}\label{app:rectified_flow_ode}

We adopt the linear interpolant process from \citet{liu2023flow}, with
\begin{equation}
    a_s = (1-s)a_0 + s\epsilon, \qquad \epsilon\sim\mathcal{N}(0,I),\quad s\in[0,1],\label{eq:interpolant}
\end{equation}
giving conditionals $p_s(a_s\mid a_0)=\mathcal{N}\left((1-s)a_0,s^2 I\right)$ with gain $G_s=1-s$ and noise scale $\sigma_s=s$.
This is the same interpolant process as in \cite{scarvelisclosed}, with the endpoint convention flipped so that $s=0$ corresponds to data and $s=1$ to noise, matching the usual diffusion convention; sampling therefore integrates from $1$ to $0$.
\citet{liu2023flow} constructs the flow ODE
\begin{equation}
    \frac{da_s}{ds} = v(a_s),
\end{equation}
where
\begin{equation}
    v(a_s) = \mathbb{E}\left[\frac{da_s}{ds} \middle| a_s\right]
\end{equation}
is the velocity term that bridges the two interpolation distributions.
\paragraph{Conditional score.}
Since $p_s(a_s\mid a_0)=\mathcal{N}\left((1-s)a_0,s^2 I\right)$, the log-density is
\begin{equation}
    \log p_s(a_s\mid a_0) = -\frac{1}{2s^2}\|a_s-(1-s)a_0\|^2 + \mathrm{const},
\end{equation}
and differentiating with respect to $a_s$ gives
\begin{equation}
    \nabla\log p_s(a_s\mid a_0) = -\frac{a_s-(1-s)a_0}{s^2}.
\end{equation}

\paragraph{Finding the velocity field.}
The minimum square-error estimate of $a_0$ given $a_s$ can be found using Tweedie's formula
\begin{equation}
    \mathbb{E}[a_0\mid a_s] = \frac{a_s+ \sigma_s^2 \nabla \log p_s(a_s)}{G_s}
    \label{eq:tweedie},
\end{equation}
or for our linear interpolant process,
\begin{equation}
    \frac{a_s + s^2\nabla\log p_s(a_s)}{1-s}.
\end{equation}

The expected velocity of the interpolant conditioned on $a_s$ can be found from differentiating Eq. \eqref{eq:interpolant}
\begin{equation}
    v_s(a_s) = \mathbb{E}\left[\frac{\mathrm{d}a_s}{\mathrm{d}s}\middle|a_s\right]
              = \mathbb{E}[-a_0+\epsilon\mid a_s]
              = \frac{a_s}{s} - \frac{1}{s}\mathbb{E}[a_0\mid a_s].
\end{equation}
The middle term here is Eq. (2) in \cite{liu2023flow}. Substituting Eq. \eqref{eq:tweedie},
\begin{gather}
    v_s(a_s) = \frac{a_s}{s} - \frac{a_s + s^2\nabla\log p_s}{s(1-s)} \\
     = \frac{-1}{1-s}\left[a_s + s\nabla\log p_s(a_s)\right].
\end{gather}
Replacing the unconditional score with the conditioned score $\nabla\log p_s(a_s\mid z)$ yields our process.

\section{Simplified Demonstrations} \label{app:simplified_demo}
\begin{wrapfigure}[13]{r}{0.35\textwidth}
\vspace{-8ex}
\includegraphics[width=\linewidth]{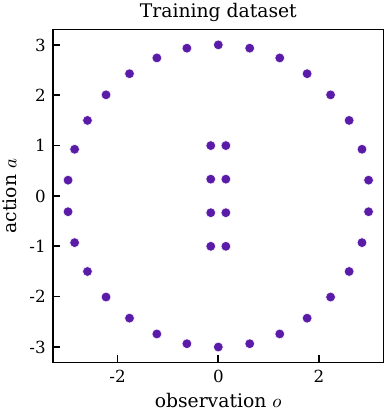}
\vspace{-2ex}\caption{Constructed dataset of 1d observations and 1d actions}
\label{fig:1d_dataset}
\end{wrapfigure}
To provide intuitions for how CFDPs perform sampling, we provide demonstrations of the inference process for a constructed dataset with 1d observation and action spaces. Figure \ref{fig:1d_dataset} shows the dataset with multimodal actions for a given observation. 

\begin{figure}[h!]
    \centering
    \includegraphics[width=\linewidth]{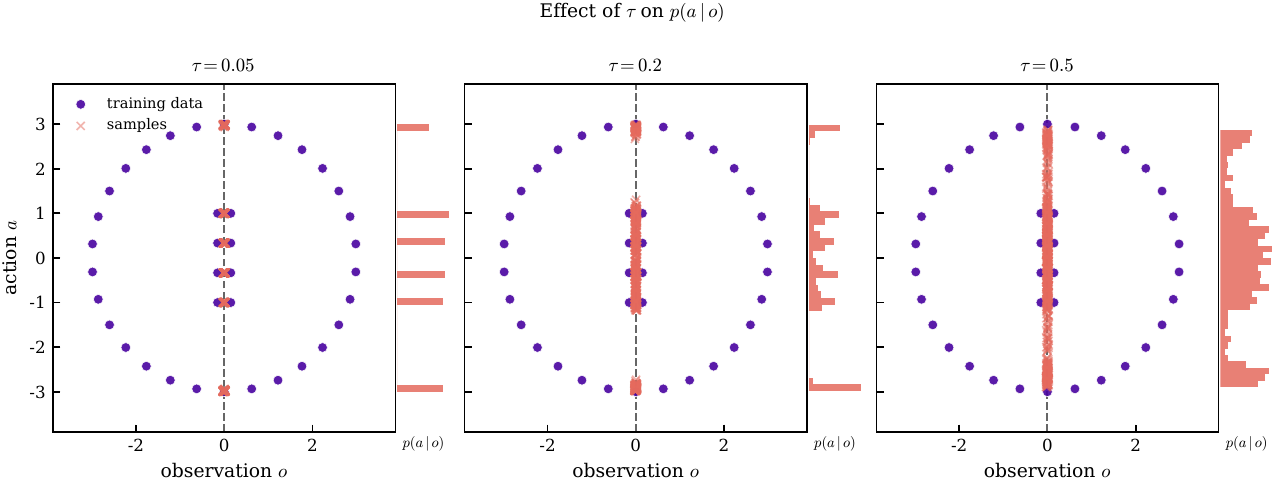}
    \caption{Inferred actions from the constructed dataset with varying levels of $\tau$, showing how the Monte Carlo smoothing controls generalization}
    \label{fig:1d_obs_action_inferences}
\end{figure}

\begin{figure}[h!]
    \centering
    \includegraphics[width=0.7\linewidth]{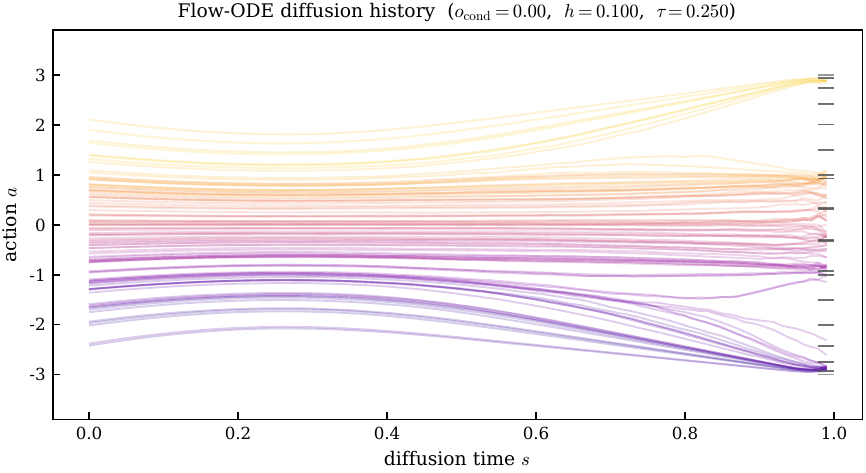}
    \caption{Flow trajectories for samples from the CFDP with 1D observation and action spaces for with $o=0$. Horizontal line segments on the right show positions of all action points in the dataset}
    \label{fig:1d_flow_evolutions}
\end{figure}

We performance inference for $o=0$ and with a CFDP model with an identity feature map, bandwidth $h=0.1$ and varying values of $\tau$. Figure \ref{fig:1d_obs_action_inferences} shows the generated actions, including histograms of their sampled distributions using CFDP with the values of $\tau$. Additionally, as there are no training samples in the middle at $o=0$, it also shows how CFDPs smooth over the observation space. 
Figure \ref{fig:1d_flow_evolutions} shows evolutions of the samples over the diffusion time for $o=0$ with the flow ODE process.

\section{Hardware Experiments}
\label{app:hardware} 
Hardware experiments were performed on a mobile class CPU on a laptop with an Intel Ultra 5 135U, 16GB of memory and no discrete GPU. All inference was done on the CPU.

The PushT $512 \text{ pixel } \times 512 \text{ pixel }$ space was mapped to a $512 \text{ mm } \times 512 \text{ mm }$ space in workspace of the UR5e arm. The arm is given 6D end effector commands using \texttt{ikpy} for inverse kinematics, with a 2D cursor position but static $z$ height and orientation of the end effector. 

The ~\PushTsymbol~ block's 6D pose is estimated by placing an ArUco fiducial marker on the table and multiple on the block. Their relative poses and a calibrated static pose between the UR5e arm and the table marker is used to map block poses of the original 2D environment poses.

The same PushT policy and parameters that are used in simulation is deployed for inference in hardware. The environment cursor positions are mapped to 6D EEF pose actions using inverse kinematics and published at 10 Hz. The EEF pose setpoints are executed by a PD controller which operates at 200 Hz and gives velocity commands to the UR5e through the \texttt{forward\_velocity\_controller ros2\_control} control interface. 

\section{Benchmark Experiments}
\label{app:experimental_details}
We implement CFDPs using the Python JAX package for implementation. We use the \texttt{hnswlib} library for fast approximate nearest neighbor search. We perform all simulated experiments on a PC with an i7-13700K, an NVIDIA RTX 3070 GPU and 32GB of memory. 
\subsection{Benchmark Evaluation}
\begin{figure}[h]
    \centering
    \includegraphics[width=0.9\linewidth]{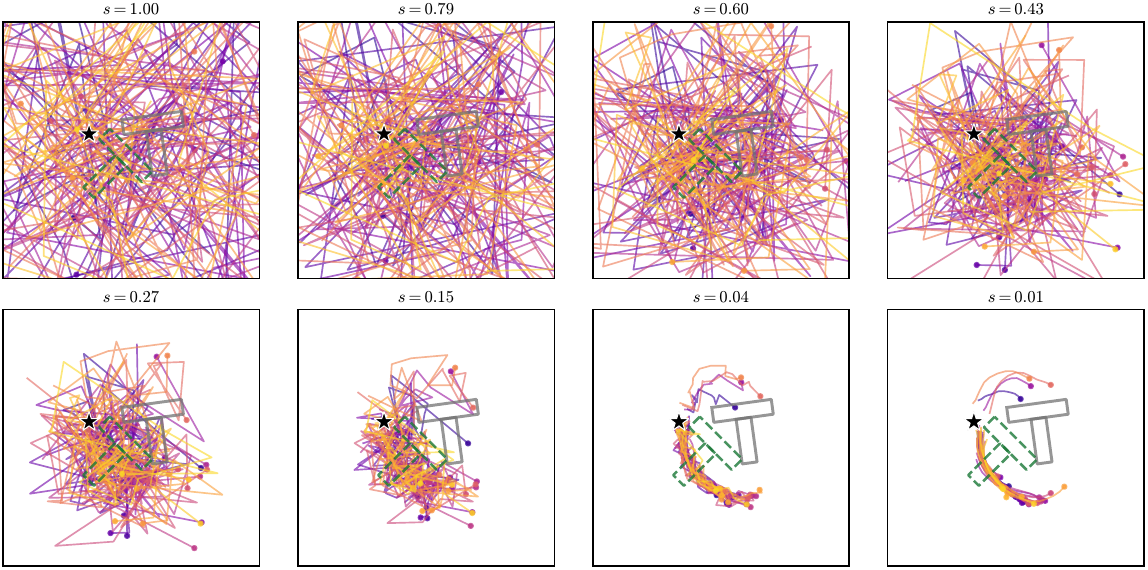}
    \caption{Trajectories inferred on the PushT environment for a flow ODE based CFDP over the diffusion time. ${\star} $ denotes the agent cursor's location.}
    \label{fig:diffusion_evolution}
\end{figure}

We evaluate CFDPs on the PushT and Robomimic tasks. We compare against results from the from Diffusion Policy~\cite{chi_diffusion_2023} paper. For fair comparison with the benchmarks, we replicate as many of the experimental details, such as number of timesteps for the eval and policy parametrizations as possible. Due to lack of a stochastic training process, the performance of CFDPs is deterministic. As a result, we compare against the checkpoint-averaged results rather than maximum performance across many checkpoints. This is also more reflective of the lack of freedom to train and evaluate many policies during data collection. Table \ref{tab:eval_env_params} provides details on the properties of the benchmark environments.
\paragraph{PushT} We use the state-based PushT dataset from \cite{chi_diffusion_2023}. We show the evolution of actions over the diffusion process in Figure \ref{fig:diffusion_evolution}. We can clearly see the multi-modality of the generated trajectories.

\paragraph{Robomimic} We use the ``low-dim'' Proficient Human (PH) datasets from the Robomimic state-based tasks ~\cite{robomimic2021}. The robot has access to end effector poses and gripper position, alongside poses of the objects being interacted with. Our results in Table \ref{tab:low-dim-results} show performance averaged over 150 episodes for each task. Similar to Diffusion Policy, we perform inference over absolute actions rather than pose deltas as stored in the original Robomimic dataset. 

\begin{table}[]
\centering
\begin{tabular}{@{}llllll@{}}
\toprule
                   & Eval time steps & \# Objects & \# Demonstrations & $D_a$ & $D_o$ \\ \midrule
\textbf{PushT}     & 300             & 1         & 200              & 2     & 5     \\ \midrule
\textbf{Lift}      & 400             & 1         & 200              & 7     & 14    \\
\textbf{Can}       & 400             & 1         & 200              & 7     & 14    \\
\textbf{Square}    & 400             & 1         & 200              & 7     & 14    \\
\textbf{Transport} & 700             & 3         & 200              & 14    & 36    \\
\textbf{ToolHang}  & 700             & 2         & 200              & 7     & 22    \\ \bottomrule
\end{tabular}
\vspace{0.5ex}
\caption{Details of benchmark environments for state-based simulation experiments}
\label{tab:eval_env_params}
\end{table}

\subsection{Dataset preprocessing}
We take demonstration dataset consisting of $N_{ep}$ episodes $\{o_{0:T}, a_{0:T}\}_0^{N_{ep}}$, and chunk them into flattened chunks with a sliding window, 
\begin{gather}
      O^{(n)}_t = \mathrm{vec}(o^{(n)}_{t - N_o + 1}, \ldots, o^{(n)}_{t}) \in \mathbb{R}^{N_o D_o}, \\
     A^{(n)}_t = \mathrm{vec}(a^{(n)}_{t}, \ldots, a^{(n)}_{t + N_a - 1}) \in \mathbb{R}^{N_a D_a}.
\end{gather}
We aggregate chunks across all episodes
\begin{gather}
    \mathcal{D} = \left\{ (O^{(n)}_t, A^{(n)}_t) : n \in [1..N_{ep}],\ \ N_o - 1 \le t \le T_n - N_a - 1 \right\}
\end{gather}
\subsection{1-NN Policy} \label{app:1nn_policy} We implement a nearest neighbor policy to have a training-free baseline. The 1-NN policy is action and observation chunked similar to CFDP, and the nearest neighbor is computed against the Mahalanobis metric to pick the action, $a_i$, associated with the observation that minimises $||o - o_i||_{\Sigma^{-1}}$ where $\Sigma = LL^\top$ as described in Appendix \ref{app:whitening}.

\subsection{Model Hyperparameters}
\label{app:experimental_params}
Here we report the parameters used for the different tasks in benchmark experiments. 

\paragraph{Bandwidth} The bandwidth, $h$, is the most important factor in determining performance. To account for how distances scale with dimensionality, we multiply by root of the (feature) dimensionality and report $h\sqrt{D_z}$.  

\begin{table}[h]
\centering
\begin{tabular}{@{}llllllll@{}}
\toprule
                   & $S$ & $k_{nn}$ & $N_a$ & $N_o$  & $\tau$ & $h\sqrt{D_z}$ \\ \midrule
\textbf{PushT}     & 100 & 2000     & 12  & 2  &  0.07    & 0.09 \\
\textbf{Lift}      & 100 & 350      & 24  & 2  &  0.02    & 0.02 \\
\textbf{Can}       & 100 & 350      & 24  & 2  &  0.05    & 0.04 \\
\textbf{Square}    & 100 & 350      & 24  & 2  &  0.05    & 0.03 \\
\textbf{Transport} & 100 & 350      & 24  & 2  &  0.06    & 0.03 \\
\textbf{ToolHang}  & 100 & 350      & 24  & 2  &  0.005   &  0.025    \\ \bottomrule \\
\end{tabular}
\caption{Hyperparameters used for Closed-Form Diffusion Policy experiments}
\vspace{0.3ex}
\label{tab:hyperparameters}
\end{table}
\newpage
\section{PushT Episode Rollouts}
\label{app:pusht_rollouts}

\begin{figure}[H]
    \centering
    \includegraphics[width=0.9\linewidth]{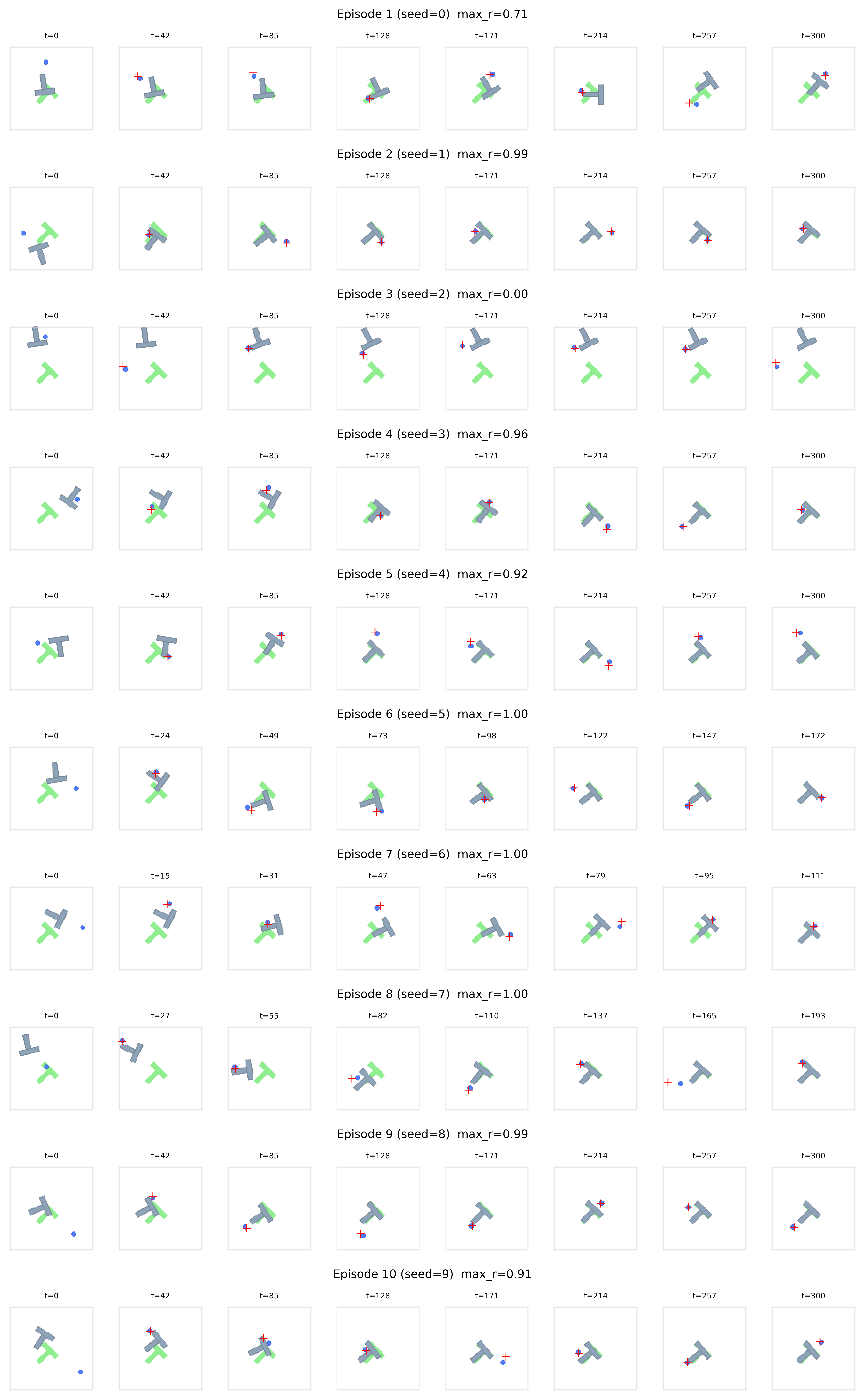}
    \caption{Example rollouts of the PushT environment with Closed-Form Diffusion Policy}
\end{figure}

\section{Neural Diffusion Policy Training}\label{app:neural_policy_training}
\begin{figure}
    \centering
    \includegraphics[width=0.70\linewidth]{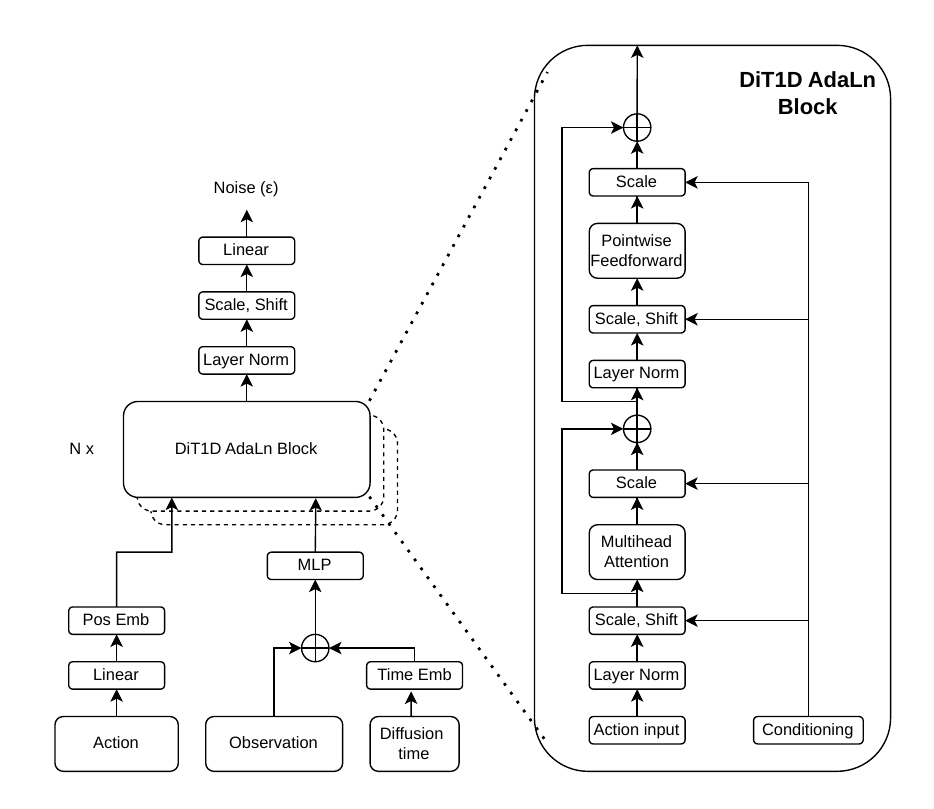}
    \caption{Architectural diagram of the variant of Diffusion Transformer used in the NDP editing experiments adapted from \texttt{clean\_diffuser}~\cite{cleandiffuser} implementation}
    \label{fig:dit_arch}
\end{figure}
We train a Diffusion Policy for the low dimensional PushT environment based on the implementation provided by \texttt{clean\_diffuser}~\cite{cleandiffuser}. This pretrained policy uses a DDPM process\cite{ho_denoising_2020}, parametrized to predict $\epsilon$, the Gaussian denoising variable in the reverse process, 
\begin{equation}
    x_{s-1} = \frac{1}{\sqrt{\alpha_s}}\left[x_s - \frac{1-\alpha_s}{\sqrt{1-\bar{\alpha}_s}}\epsilon_{\theta}(x_s,s) \right] + \varsigma_s z_s \label{eq:ddpm},
\end{equation}
where $\alpha$, $\bar \alpha$ and $\varsigma$ are parameters that depend on the noise schedule. The diffusion backbone uses a 1D variant of Diffusion Transformer (DiT) architecture with architecture shown in Figure \ref{fig:dit_arch}, with a total of 4.07 million parameters, 10 attention heads, with two sequential DiT blocks. The DiT architecture is conditioned by an MLP with 256 hidden units and 128 dimension output embedding. The diffusion backbone use a Cosine noise schedule for $\beta_s = 1-\alpha_s$, and $100$ diffusion time steps. The model is trained for $10^6$ gradient steps, with a batch size of $256$. The training process takes 4 hours and 20 minutes on an RTX 3070.

\section{Neural Policy editing}\label{app:neural_editing} 
\subsection{Training-free Classifier-Free Scores}
\label{app:policy_edit_guidance}

\begin{figure}
    \centering
    \includegraphics[width=0.8\linewidth]{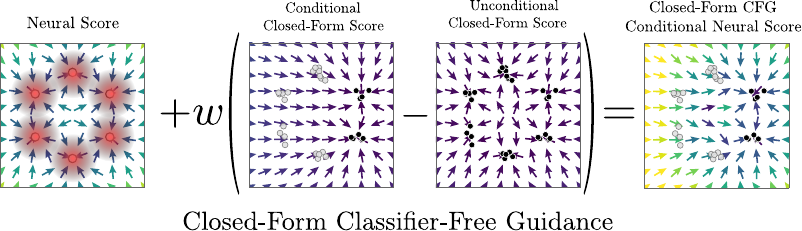}
    \caption{Conceptual figure showing how closed-form scores can be used to guide the neural diffusion policies. In this example, the goal is to sample points specifically from the two right-most modes}
    \label{fig:cfg_conceptual}
\end{figure}
Given a partitioning of $\mathcal{A}$, into a categorical distribution, $\mathcal{C}$, we can sample $p(a \mid o, c)$, with $c \in \mathcal{C}$, using an NDP trained to sample from $p(a \mid o)$ and data labeled match condition $c$. We follow the usual derivation for classifier-free guidance. For conditional value $c \in \mathcal{C}$, we expand the the target $\mathcal{C}$-conditioned distribution using Bayes' rule,
\begin{equation}
    p_s(a \mid o, c) = \frac{p_s(a \mid o) p_s(c \mid a, o)}{p_s(c\mid o)}.
\end{equation}
We take the log,
\begin{equation}
    \log p_s(a \mid o, c) = \log p_s(a \mid o) + \log p_s(c \mid a, o) - \log p_s(c\mid o),
\end{equation}
and then take the gradient with respect to $a$ to find the score, which eliminates $\log p_s(c\mid o)$
\begin{align}
    \nabla \log p_s(a \mid o, c) &= \nabla(\log p_s(a \mid o) + \log p_s(c \mid a, o) - \log p_s(c\mid o))  \\
    &= \underbrace{\nabla \log p_s(c \mid a, o )}_{\text{Classifier likelihood}} +  \underbrace{\nabla \log p_s(a \mid o )}_{\mathcal{C}-\text{unconditional score}}.
\end{align}
Following classifier-free guidance, to avoid training a noise-conditioned classifier, we express the classifier term in terms of conditional and unconditional scores that we can obtain from two CFDPs.
\begin{equation}
    \nabla \log p_s(a \mid o, c) = \nabla \log p_s(a \mid o) + w[\nabla \log p_s(a\mid o, c) - \nabla \log p_s(a \mid o)]
\end{equation}
Lastly, we take the first term from the NDP and the guidance term from the closed-form score.

\subsection{Classifier-Free Guidance Implementation}
\label{app:cfdp_cfg_details}

To perform policy guidance, we take the pretrained $\epsilon$-predicting network and convert to a score and run a score-based version of the DDPM process,
\begin{equation}
    x_{s-1} = \frac{1}{\sqrt{\alpha_s}}\left[x_s + (1-{\alpha}_s) \nabla \log p_s(x_s) \right] + \varsigma_s z_s \label{eq:ddpm2}.
\end{equation}
We convert the pre-trained $\epsilon$-predicting neural network into a score prediction using
\begin{equation}
    \nabla \log p_s(x_s) = \frac{1}{\sqrt{1 - \bar \alpha_s}}\epsilon_\theta(x_s, s)
\end{equation}

In theory, for samples matching the desired condition, conditional score realigns with the unconditional score after class-specific symmetry breaking has occurred~\cite{jin_stage-wise_2025}, making the guidance vector approach $\vec 0$, and thus the neural diffusion model dominates at the later denoising stages. This is also apparent in equation \eqref{eq:CF-Score} as near the condition-matching data points, the unconditional score's softmax input is dominated by nearby points.

In practice, we don't apply guidance for the last few timesteps of the policy, because while the guidance vector should vanish near condition matching samples, the neural policy may not have a $c$-matching mode actions in its conditional distribution. This means at later timesteps, the guidance score might be very large. Meanwhile, for a policy which has a $c$ condition meeting mode, we only need to put the diffusion state, $a_s$, in the basin of attraction of that mode, so lack of guidance in later timesteps should be sufficient. We then implement guidance using Equation \eqref{eq:cfdp_cfg}. Table \ref{tab:policy_edit_params} shows the parameters used for our Closed-Form Diffusion Policy and the Neural Diffusion Policy.
\begin{table}[H]
    \centering
\begin{tabular}{@{}llllllll@{}}
\toprule
                   & $S$ & $k_{nn}$ & $H$ & $N_o$ & $N_a$ & $\tau$ & $h$ \\ \midrule
\textbf{CFDPs}     & 100 & 2000     & 8   & 2     & 8     & 0.07     & 0.09    \\
\textbf{DP-T}     & 100 & N/A       & 8   & 2     & 8     & N/A      & N/A  \\
\bottomrule \\
\end{tabular}
\caption{Parameters used by the conditional and unconditional CFDPs, and the Neural Transformer-based Diffusion Policy}
\label{tab:policy_edit_params}
\end{table}

\begin{algorithm}[h]
    \label{algo:cfdp_cfg}
    \caption{Neural Diffusion Policy with Closed-Form Guidance}
    
    \KwIn{Labeled Dataset $\mathcal{D} = \{\hat a_i, \hat o_i, \hat c_i\}_0^N$, CFDP parameters ($h, \tau, k_{nn}$, $N_a$, $N_o$), Feature map $\phi$, Neural policy $\psi_\theta$, Guidance strength $w$, Target condition $c$, Time threshold $s_{\text{guidance}}$}
    Construct k-NN tree ($\mathcal{T}$) from dataset ($\mathcal{D}$)
    
    Construct conditional k-NN tree ($\mathcal{T}_c$) from filtered dataset ($\{\mathcal{D} :\hat c_i = c\}$)

    \While{$t \gets t + H$}{
        Sample $a_S \sim \mathcal{N}(0, I)$ \\
        $o \gets $ Sense and chunk observation history from robot \\
        Map observation $z \gets \phi(o)$ \\
        Query $\underline{\mathcal{D}}$ and $\underline{\mathcal{D}}_c$ from $\mathcal{T}$ and $\mathcal{T}_c$\\ 
        \For{$s \gets$ $S..1$ steps}{
            Compute $\nabla \log p^{CF}_s(a_s \mid z)$, and $\nabla \log p^{CF}_s(a_s \mid z,\, c)$ from $\underline{\mathcal{D}}$ and $\underline{\mathcal{D}}_c$ via \eqref{eq:smoothed_cond_kernel}

            Get score from neural policy $\nabla \log p^{NN}_s(a_s \mid z) \approx \psi_\theta(a_s, s)$

            \eIf{$s > s_{\text{guidance}}$}{
                Find guided score via \eqref{eq:cfdp_cfg}
            }{
                Use neural score
            }

            $a_{s-1} \gets $ according to diffusion step \eqref{eq:ddpm2}
        }
        $u_{t:t+H} \gets a_0$ \\
        Execute actions on robot for $H$ timesteps
    }
\end{algorithm}

\subsection{Novel Demonstration Augmentation} \label{app:augmentation}
A neural diffusion policy trained on a separate corpus $\mathcal{D}^{\mathrm{NN}}$ implicitly represents a conditional $p_\theta(a\mid o)$ over a large number of observations. We would like to inherit this property without retaining $\mathcal{D}^{\mathrm{NN}}$ itself. We assume that instead of retaining the entire training dataset, we only maintain basic statistics of $\mathcal{D}^{\mathrm{NN}}$ such as the mean observation and the empirical covariance, $\Sigma^{NN}$.

We therefore summarize the entire neural contribution by a \emph{single} virtual data point whose location changes at every reverse step. Concretely, at diffusion step $s$ we use the $\epsilon$-prediction from the Neural Diffusion Policy to predict a final output using Tweedie's formula for a DDPM process,
\begin{equation}
\hat a_0^{\mathrm{NN}} = \frac{1}{\sqrt{\bar\alpha_s}}\left(a_s - \sqrt{1-\bar\alpha_s}\,
\epsilon_\theta(a_s, s)\right).
\end{equation}
We treat $\hat a_0^{\mathrm{NN}}$ as an additional virtual data point in the score computation. The data point is centred at the empirical mean $\mu_o^{\mathrm{NN}}$ of the neural training observations, with the kernel metric using the empirical covariance $\Sigma_o^{\mathrm{NN}}$, scaled by a factor $c$ that is a hyperparameter that trades off the effect of the neural and the augmenting data points.

\paragraph{Augmented kernel.}
We now express our original joint distribution as a heterogeneous mixture 
\begin{equation}
    p(a, z) \propto \underbrace{\delta\left(a-\hat a_0^{NN}\right) K_{h}^{NN}\left(z-\mu_z^{NN}\right)}_{\text{Neural virtual data point}} + \sum_{i=1}^{K}\underbrace{\delta(a-\hat a_i)K_{h}\left(z-\hat z_i\right)}_{\text{Augmenting demonstrations}}.
    \label{eq:augmented_joint}
\end{equation}
We extend the previous derivation of Equation \eqref{eq:joint_kde} and rewrite the joint with a data dependent  $K^{(i)}_h(\cdot)$  that is non-isotropic Gaussian kernel,
\begin{equation}
    K^{(i)}_h(x) = (2\pi)^{-D_z/2} (\det H^{-1}_i)^{-1/2} \exp\left(-\frac{1}{2}x^TH_i^{-1}x\right).
\end{equation}
This non-isotropic data dependent kernel provides our neural data point with scaled-covariance influence when ($H_0 = c\Sigma^{NN}$). In addition, this can also subsume our Mahalanobis kernel as it provides non-isotropic influence through the non-isotropic local covariance. We use $c=0.8$ for our experiments.

We recall Equation \eqref{eq:weights_raw},  which provided the weight of each data point's contribution to the score computation and make it datapoint dependent
\begin{equation}
    w_i = \frac{(2\pi\sigma^2)^{-D_a/2}\exp\left(-\|a_s - G \hat a_i\|^2/(2\sigma^2)\right)K^{(i)}_h(z - \hat z_i)}
                    {\sum_j (2\pi\sigma^2)^{-D_a/2}\exp\left(-\|a_s - G a_j\|^2/(2\sigma^2)\right)K^{(j)}_h(z - z_j)}.
\end{equation}
From here on, the only structural change from the homogeneous derivation is that each data point's observation kernel carries its own covariance $H_i$, and as data point uses different covariances, the softmax retains a data-dependent normalizer term. This leaves the data 
\begin{equation}
    \log w_i^{\mathrm{obs}} = -\tfrac{1}{2}(z-\mu_i)^{\top} H_i^{-1}(z-\mu_i) - \tfrac{1}{2}\log\det H_i,
    \quad
    (\mu_i, H_i) =
    \begin{cases}
        (\mu_z^{NN}, c\Sigma^{\mathrm{NN}}) & i = \text{neural},\\[2pt]
        (\hat z_i, h^2 I) & i = 1,\dots,K.
    \end{cases}
    \label{eq:augmented_obs_weight}
\end{equation}
This is a generalization of both NDPs and CFDPs: If there is no extra data collected, only the neural data point remains and the softmax collapses and the augmented sampler reduces exactly to the pre-trained neural DDPM reverse step. If no neural data is collected, this is equivalent to the original CFDP weight \eqref{eq:weights_softmax}. In our implementation, the network is queried only once per diffusion step and $\hat a_0^{NN}$ is held fixed across the $M$ perturbations and the smoothing affects the CFDP demonstration entries only. The reverse update then uses the score-based DDPM step \eqref{eq:ddpm2}.

\begin{algorithm}[h]
    \label{algo:cfdp_augmentation}
    \caption{Closed-Form Novel Demonstration Augmentation}

    \KwIn{Augmentation Dataset $\mathcal{D} = \{\hat a_i, \hat o_i\}_0^N$, CFDP parameters ($h, \tau, k_{nn}$, $N_a$, $N_o$), Feature map $\phi$, Neural policy $\psi_\theta$, Empirical neural dataset observation mean, $\mu_o^{NN}$ and covariance $\Sigma_o^{NN}$}
    Construct k-NN tree ($\mathcal{T}$) from dataset ($\mathcal{D}$)

    \While{$t \gets t + H$}{
        Sample $a_S \sim \mathcal{N}(0, I)$ \\
        $o \gets $ Sense and chunk observation history from robot \\
        Map observation $z \gets \phi(o)$ \\
        Query $\underline{\mathcal{D}}$ from $\mathcal{T}$\\ 
        \For{$s \gets$ $S..1$ steps}{
            Get score from neural policy $\nabla \log p^{NN}_s(a_s \mid z) \approx \psi_\theta(a_s, s)$

            Use Tweedie's formula to predict $\hat a^{NN}_0 = \mathbb{E}[a_0|a_s]$ via \eqref{eq:tweedie}

            Create augmented dataset  $\mathcal{D}_{s}^{\text{aug}} = \underline{\mathcal{D}} \cup \{(\mathbb{E}[a_0|a_s],\, \mu_o^{NN})\}$
        
            Compute augmented score $\nabla \log p_s(a_s\mid z)$ from $\mathcal{D}_{s}^{\text{aug}}$ according to \eqref{eq:augmented_kernel} and \eqref{eq:augmented_obs_weight}

            $a_{s-1} \gets $ according to diffusion step \eqref{eq:ddpm2}
        }
        $u_{t:t+H} \gets a_0$ \\
        Execute actions on robot for $H$ timesteps
    }
\end{algorithm}
\subsection{Dataset Partitioning}
We programmatically generate our labeled datasets but note that CFDP did not have access to the original labeling strategy and that datasets could be human labeled. 
\paragraph{Policy Guidance} The objective of the rotational labeling procedure is to isolate final orientation adjustments and assign each one a clockwise or counter-clockwise label. To label demonstrations, we first sweep through each demonstration and look for frames where the cursor is near the head of the ~\PushTsymbol~ block while the block is within 30 pixels of the end position. Once a frame has been identified, we calculate the change in angle of the cursor from the center of the screen over the next 30 timesteps and classify it as clockwise if the angle moves clockwise by more than 0.25 radians. 

The conditional dataset contains all timesteps except those that were found to not contain clockwise movement after detection was triggered. Closed-form scores were constructed from both the original dataset and this conditional dataset.
\paragraph{Demonstration Augmentation} The goal was to show that CFDP could be used to patch two different policies trained on disjoint datasets. The datasets are labeled as left or right entirely based on whether the centre of the block is on the left half $(x < 256)$ or the right half $(x > 256)$. This roughly splits half of the data out of each dataset. 

\section{Hyperparameter studies}\label{sec:effect_of_hyperparams}
We perform studies to show the effect of CFDP hyperparameters on performance. We use Robomimic's Square task as a test case to study the behavior of CFDP. We conduct logarithmic scale sweeps for $\tau$ and $h$, with performance being evaluated for each hyperparameter value over 150 episodes. All other values are kept the same as the hyperparameters for the Square task in Appendix \ref{app:experimental_params}. Figure \ref{fig:h_tau_sweeps} show large bands of acceptable performance where the performance is consistent across orders of magnitude variations, peaks where performance is higher before performance drops off due to oversmoothing. 
\begin{figure}[h]
    \centering
    \includegraphics[width=0.48\linewidth]{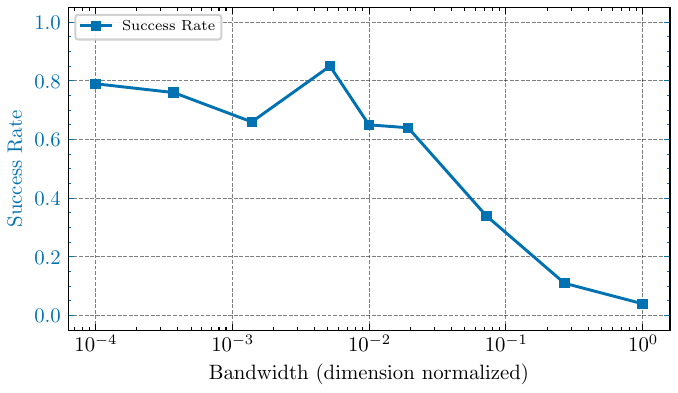}
    \includegraphics[width=0.48\linewidth]{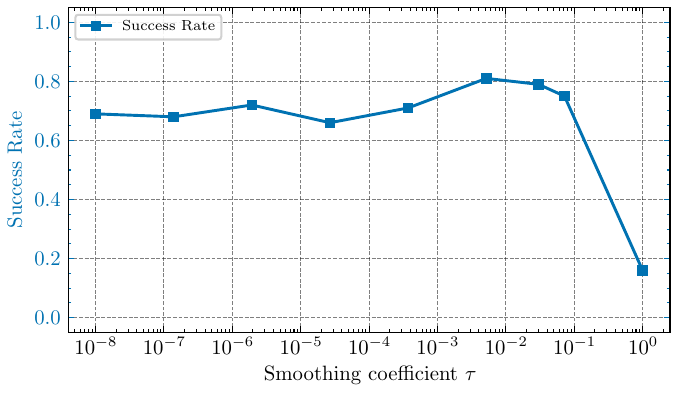}
    \caption{Effect of $h\sqrt{D_z}$ and $\tau$ on performance}
    \label{fig:h_tau_sweeps}
\end{figure}

\begin{figure}[h]
    \centering
    \includegraphics[width=0.48\linewidth]{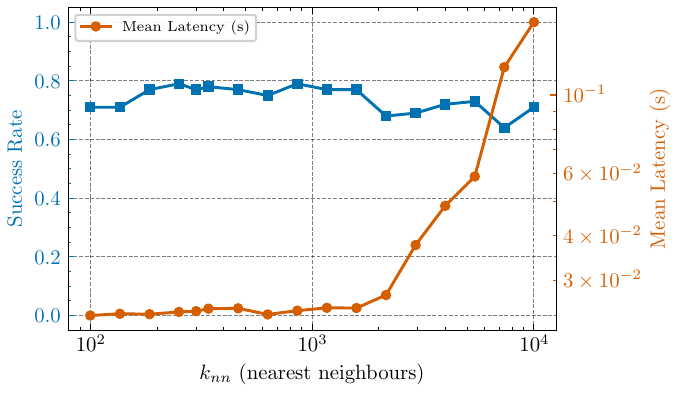}
    \includegraphics[width=0.48\linewidth]{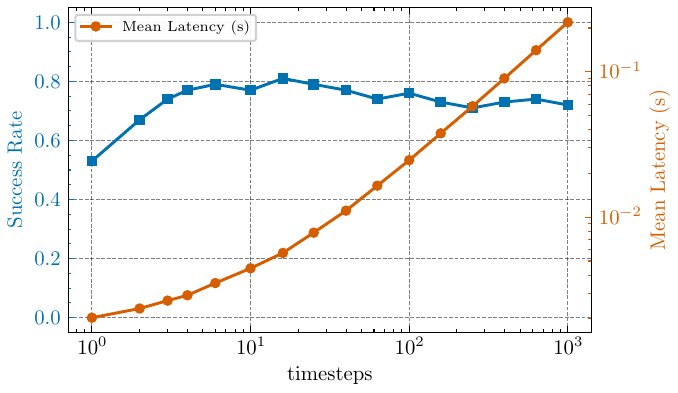}
    \caption{Performance and inference latency as a function of $k_{nn}$ neighbors and $S$ timesteps for the Robomimic Square task. }
    \label{fig:latency_plots}
\end{figure}

We also conduct sweeps of $k_{nn}$ and timesteps, $S$, both parameters which heavily affect inference latency and can be reduced for faster inference. Figure \ref{fig:latency_plots} shows how various values of $k_{nn}$ and $S$ still maintain high performance at the task at lower values. Both too low and too high values of $k_{nn}$ reduce performance, as it controls geometry of smoothing in the observation space. 

\clearpage

\end{document}